\documentclass[11pt]{article}
\usepackage{makecell}
\usepackage[preprint]{acl}
\usepackage{booktabs}
\usepackage{amsmath}
\usepackage{mathtools}
\usepackage{multirow}
\usepackage{caption} 
\usepackage{times}
\usepackage{latexsym}
\usepackage[table]{xcolor}
\usepackage[breakable,skins]{tcolorbox}

\newcommand*\justify{%
  \fontdimen2\font=0.4em
  \fontdimen3\font=0.2em
  \fontdimen4\font=0.1em
  \fontdimen7\font=0.1em
  \hyphenchar\font=`\-
}

\renewcommand{\texttt}[1]{%
\begingroup
\ttfamily
\begingroup\lccode`~=`/\lowercase{\endgroup\def~}{/\discretionary{}{}{}}%
\begingroup\lccode`~=`[\lowercase{\endgroup\def~}{[\discretionary{}{}{}}%
\begingroup\lccode`~=`.\lowercase{\endgroup\def~}{.\discretionary{}{}{}}%
\catcode`/=\active\catcode`[=\active\catcode`.=\active
\justify\scantokens{#1\noexpand}%
\endgroup
}

\PassOptionsToPackage{numbers, compress}{natbib}

\usepackage[utf8]{inputenc}         %
\usepackage[T1]{fontenc}            %
\usepackage{graphicx}
\usepackage{todonotes}
\usepackage{multirow}
\usepackage{hyperref}
\usepackage[capitalize,noabbrev,nameinlink,sort]{cleveref}
\usepackage{subcaption}
\usepackage{multicol}
\usepackage{array}
\usepackage{enumitem}
\usepackage{algorithm}
\usepackage{algpseudocode}
\usepackage{tabularx}
\usepackage{listings}
\usepackage{makecell}
\usepackage{xspace}
\usepackage{colortbl}
\usepackage{fontawesome5}
\usepackage{pifont}
\usepackage[normalem]{ulem}
\useunder{\uline}{\ul}{}
\usepackage{tablefootnote}
\usepackage{tcolorbox}
\usepackage{arydshln}
\usepackage[toc, page, header]{appendix}
\usepackage{tabularx}
\usepackage{listings}

\usepackage{amsmath}
\usepackage{amssymb}
\usepackage{amsfonts}                               
\usepackage{amsthm}
\usepackage[mathcal]{eucal}
\usepackage{mathrsfs}
\usepackage{bm}                                     
\usepackage{blkarray}                               
\usepackage{nicefrac}                               
\usepackage{bbm}

\usepackage{wrapfig}
\usepackage{graphicx}                               
\usepackage{caption}
\usepackage{tikz}
\usepackage{circuitikz}
\usetikzlibrary{patterns,snakes}
\usetikzlibrary{positioning,calc,fit,decorations.pathmorphing,shapes.geometric, shapes.gates.logic.US, calc}
\usetikzlibrary{arrows,arrows.meta,decorations.markings,shapes,shapes.arrows}
\usetikzlibrary{decorations,decorations.pathreplacing}
\usetikzlibrary{backgrounds}
\usepackage{filecontents}                           
\usepackage{pgfplots}
\usepackage{pgfplotstable}
\usepgfplotslibrary{groupplots}
\usepackage{scalefnt}
\pgfplotsset{compat=newest}
\usepackage{xcolor}

\definecolor{firstcolor}{HTML}{C3423F}
\definecolor{secondcolor}{HTML}{2A4B8C}
\definecolor{aworld_blue}{HTML}{4e81ff}
\definecolor{aworld_cyan}{HTML}{41d7fa}
\definecolor{aworld_teal}{HTML}{5fede4}
\definecolor{coral}{RGB}{255,127,80}
\definecolor{darkgreen}{RGB}{0,100,0}
\definecolor{darkyellow}{RGB}{204,153,0}
\definecolor{salmon}{RGB}{250,128,114}
\definecolor{darkred}{RGB}{150,0,0}

\hypersetup{
  colorlinks=true,
  linkbordercolor=aworld_blue,
  citebordercolor=aworld_cyan,
  urlbordercolor=aworld_teal
}

\definecolor{improvementblue}{RGB}{55,126,184}    
\definecolor{degradationorange}{RGB}{230,85,13}   

\def\eqref#1{equation~\ref{#1}}

\def\1{\bm{1}}

\DeclareMathAlphabet{\mathsfit}{\encodingdefault}{\sfdefault}{m}{sl}
\SetMathAlphabet{\mathsfit}{bold}{\encodingdefault}{\sfdefault}{bx}{n}

\usepackage[T1]{fontenc}
\usepackage{pifont}
\usepackage{bbm}

\usepackage[utf8]{inputenc}
\usepackage{subcaption}
\usepackage{microtype}
\usepackage{bbding}
\usepackage{inconsolata}
\usepackage{graphicx}

\title{}
\author{}

\begin{document}

\twocolumn[{%
\noindent\includegraphics[height=1.0cm]{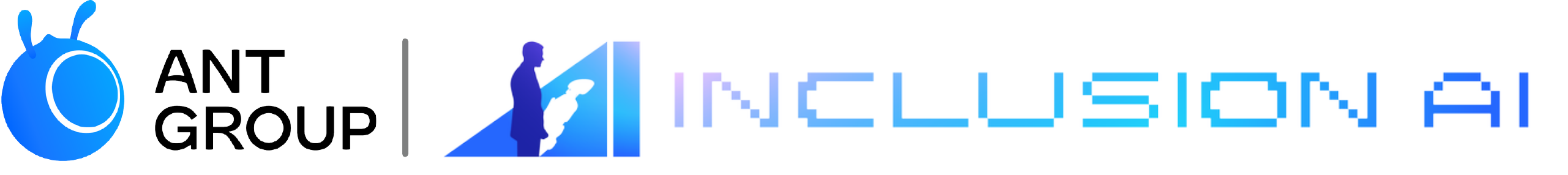}\\[0.3em]
\noindent\rule{\textwidth}{0.4pt}

\vspace{0.5em}
\begin{center}
{\Large\textbf{V2P: Visual Attention Calibration for GUI Grounding via Background Suppression and Center Peaking}}\\[0.2em]
\end{center}
\vspace{0.4em}

\noindent\rule{\textwidth}{0.4pt}

\vspace{0.6em}
\begin{center}
\textbf{Jikai Chen\textsuperscript{1,2*}}, \textbf{Long Chen\textsuperscript{2*}}, \textbf{Dong Wang\textsuperscript{2*}},
\\[0.3em]
\textbf{Qinglin Su\textsuperscript{1}}, \textbf{Zhixuan Chu\textsuperscript{1}}, \textbf{Bingguang Hao\textsuperscript{2}}, \textbf{Leilei Gan\textsuperscript{1$\ddag$}},
\\[0.3em]
\textbf{Chenyi Zhuang\textsuperscript{2$\ddag$}}, \textbf{Jinjie Gu\textsuperscript{2}}
\\[0.5em]
\small
\textsuperscript{1}Zhejiang University \quad
\textsuperscript{2}Inclusion AI, Ant Group
\\[0.3em]
\raisebox{-0.1em}{\includegraphics[height=0.9em]{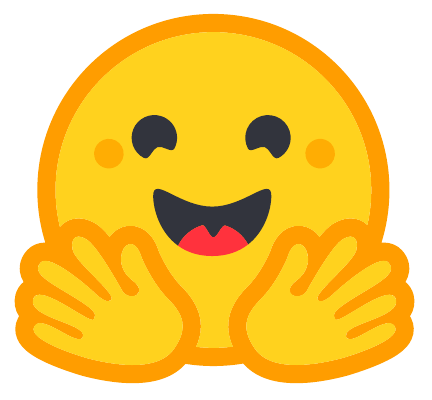}} \href{https://huggingface.co/Minstrel54524/V2P-7B}{\texttt{Model}}\quad\quad\faGithub\ \href{https://github.com/inclusionAI/AgenticLearning/tree/main/V2P}{\texttt{Code}}
\end{center}

\vspace{0.8em}

\noindent
\begin{minipage}{\textwidth}
  \centering
  \begin{minipage}{0.48\textwidth}
    \centering
    \includegraphics[width=\linewidth]{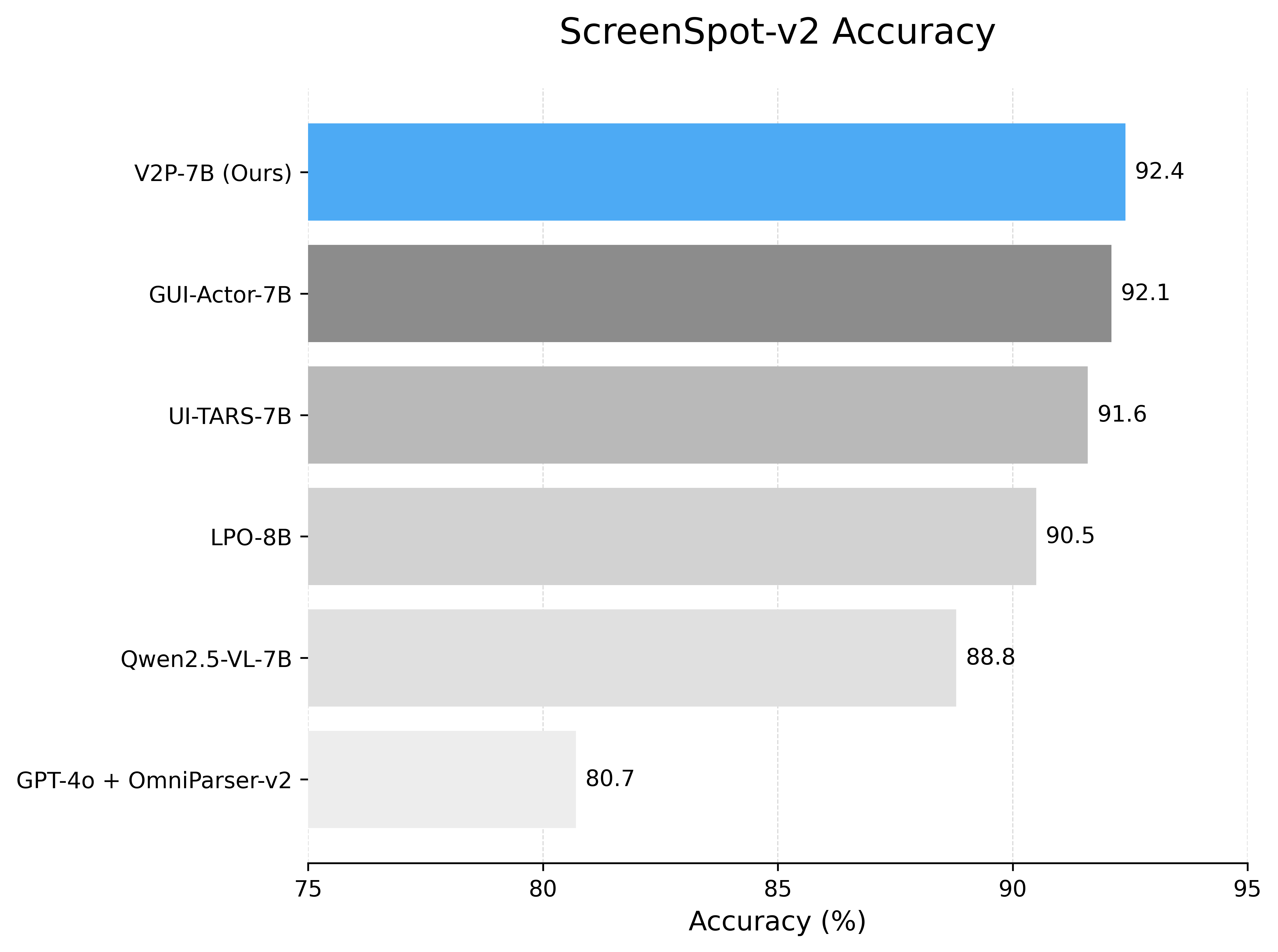}\\[0.3em]
    \small (a) ScreenSpot-v2
    \label{fig:screenspot_v2_chart}
  \end{minipage}%
  \hfill
  \begin{minipage}{0.48\textwidth}
    \centering
    \includegraphics[width=\linewidth]{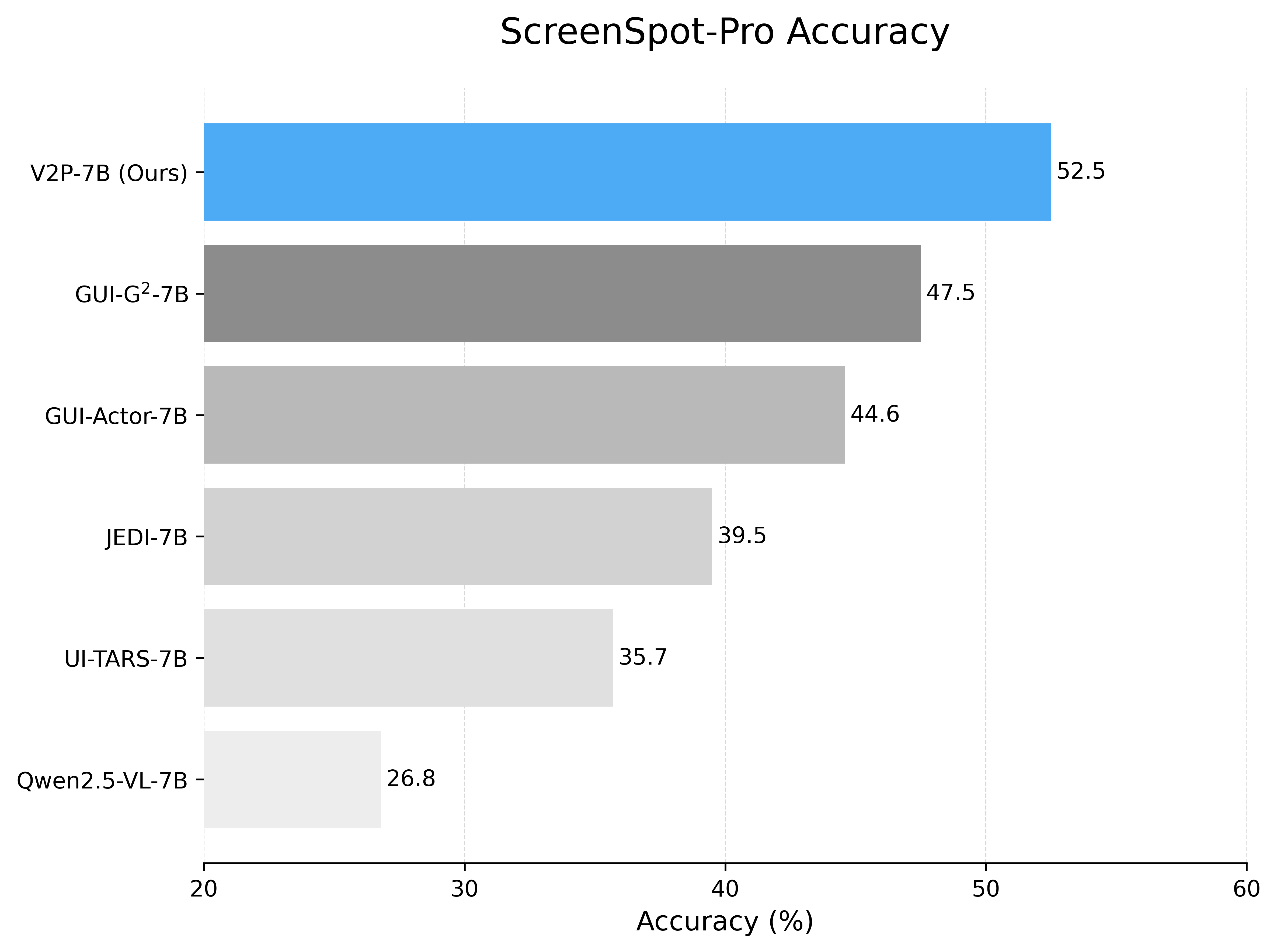}\\[0.3em]
    \small (b) ScreenSpot-Pro
    \label{fig:screenspot_pro_chart}
  \end{minipage}
  \vspace{0.3em}
  \begin{center}
  \small Performance comparison of different baselines and our V2P-7B model on ScreenSpot-v2 (left) and ScreenSpot-Pro (right). By training Qwen2.5-VL-7B using our Valley-to-Peak training strategy, our model achieves the best performance among all competitors.
  \phantomsection\label{fig:main_results_charts}
  \end{center}
\end{minipage}

\vspace{0.5em}
}]

{\renewcommand\thefootnote{}
\footnotetext{*Equal contributions. $\ddag$Corresponding Authors.}}

\begin{abstract}
Precise localization of GUI elements is crucial for the development of GUI agents. Traditional methods rely on bounding box or center-point regression, neglecting spatial interaction uncertainty and visual-semantic hierarchies. Recent methods incorporate attention mechanisms but still face two key issues: (1) ignoring processing background regions causes attention drift from the desired area, and (2) uniform modeling the target UI element fails to distinguish between its center and edges, leading to click imprecision. Inspired by how humans visually process and interact with GUI elements, we propose the Valley-to-Peak (V2P) method to address these issues. To mitigate background distractions, V2P introduces a suppression attention mechanism that minimizes the model's focus on irrelevant regions to highlight the intended region. For the issue of center-edge distinction, V2P applies a Fitts' Law-inspired approach by modeling GUI interactions as 2D Gaussian heatmaps where the weight gradually decreases from the center towards the edges. The weight distribution follows a Gaussian function, with the variance determined by the target's size. Consequently, V2P effectively isolates the target area and teaches the model to concentrate on the most essential point of the UI element. The model trained by V2P achieves the performance with 92.4\% and 52.5\% on two benchmarks ScreenSpot-v2 and ScreenSpot-Pro. Ablations further confirm each component's contribution, underscoring V2P's generalizability in precise GUI grounding tasks and its potential for real-world deployment in future GUI agents.
\end{abstract}
\section{Introduction}

Recent advances in large language models (LLMs) and vision-language models (VLMs) have enabled agents to interpret natural language instructions and interact with graphical user interfaces (GUIs) across desktop, mobile, and web platforms. Central to this capability is GUI grounding, which aligns language commands with semantically relevant UI elements and their spatial locations~\citep{seeclick}. This task bridges user intent and interface actions, supporting the development of intelligent, general-purpose agents for real-world human-computer interaction.

Early approaches framed GUI grounding as coordinate generation task, outputting a bounding box or $(x, y)$ coordinate for a natural-language query~\citep{gui_survey, ui-tars}. However, this “coordinate generation” method suffers weak spatial--semantic alignment~\citep{gui-actor}, treating coordinates like ordinary words without inherent spatial meaning. Moreover, point-wise regression contradicts the multi-point validity inherent in real interactions. Recent work addresses these issues by leveraging the model’s attention maps~\citep{gui-actor}. Instead of predicting coordinates, it extracts cross-modal attention weights linking instruction tokens to image patches, selecting the most attended patch as the click position. This approach offers dense spatial supervision and naturally tolerates multiple valid click regions, aligning better with human behavior.

\begin{figure*}[t]
    \centering
    \includegraphics[width=0.9\linewidth]{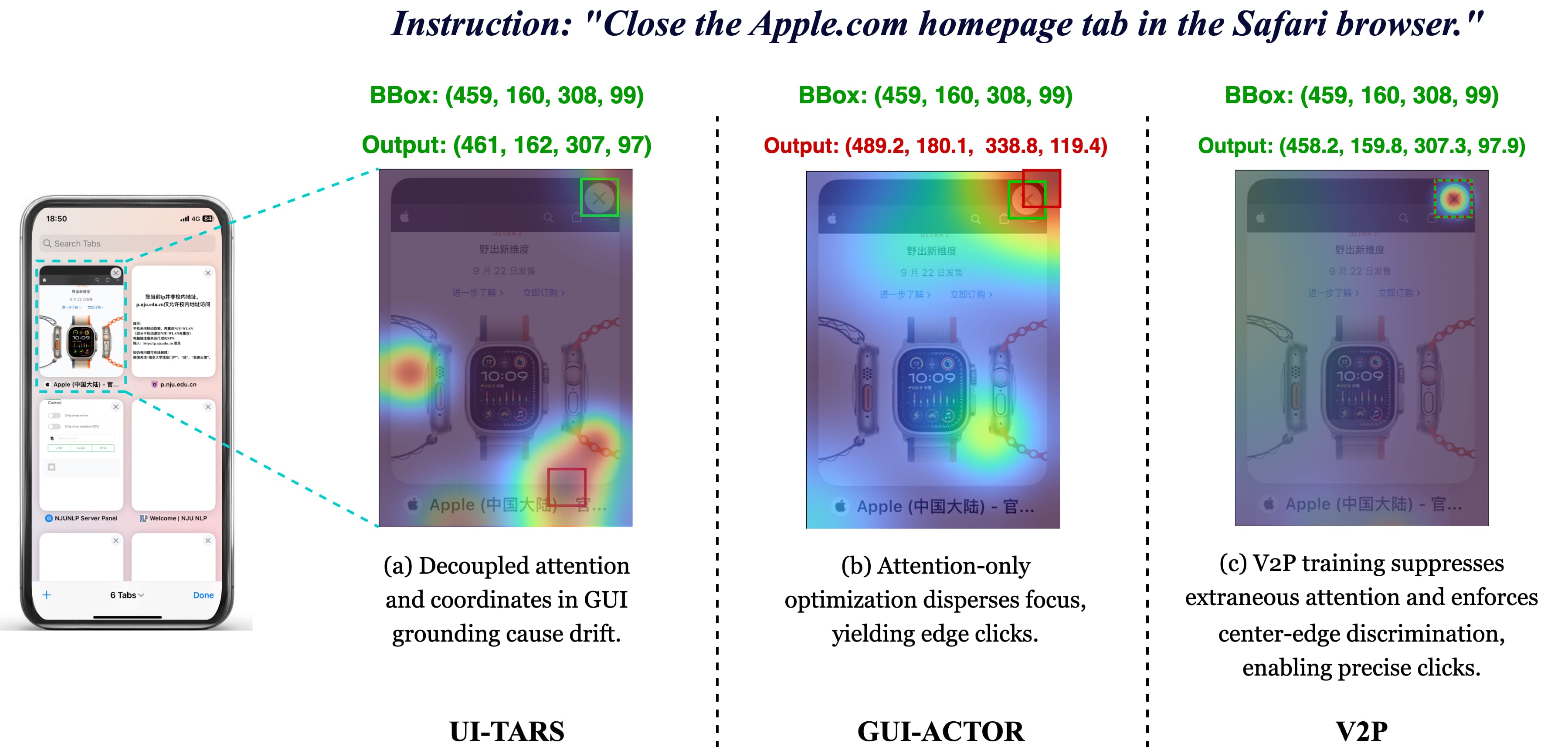}
    \caption{Comparison of different strategies in the GUI grounding task. The green box marks the ground-truth bounding box, and the red box highlights the region where the model places the highest attention given the instruction and screenshot. The overlaid heatmap is colour-coded from cool (blue) to warm (red), with warmer colours indicating higher attention values.}
    \label{fig:illustration}
\end{figure*}

However, after manually scrutinizing the attention heatmap of these methods mentioned above, we found two main issues, as shown in Fig.~\ref{fig:illustration}:
\begin{enumerate}
    \item \textbf{Background Distraction}: Current loss functions only reward attention on target patches but fail to penalize it on the background. This leads to a "divergent" attention distribution where background regions also receive high scores. Consequently, softmax normalization allows these regions to absorb probability mass, weakening or even shifting the intended attention peak.
    \item \textbf{Centre-edge Confusion}: Because labels treat all pixels within a bounding box equally, the model cannot differentiate an element's center from its edges, resulting in uniform attention and inaccurate clicks that miss the center. Furthermore, for small elements, this often leads the attention to drift towards the edges, making the model more prone to mislocalization, especially when elements overlap.
\end{enumerate}

This raises a key question: \emph{How can we guide the model’s attention to focus more precisely on the target UI element?} Motivated by human behavior—first isolating the target (valley suppression) then focusing on the action point (peak emphasis)—we propose \textbf{Valley-to-Peak (V2P)}. V2P suppresses distractions by creating low-attention "valleys" in irrelevant areas while sharpening a "peak" at the actionable center.

\textbf{Suppression Attention:} We apply inverse attention regularization~\citep{suppression-attention} to penalize high attention outside the target, isolating true UI elements and reducing attention to non-target regions.

\textbf{Fitts-Gaussian Peak Modeling:} Inspired by Fitts’ Law~\citep{gauss_1, gauss_2}, we use a 2D Gaussian centered on the target, scaled to its size, to model human's click likelihood, which yields a heatmap that peaks at the center and decays towards the edges, better matching real user interactions. 

Together, these modules reshape the attention map, enhancing grounding precision by aligning the model’s focus with human patterns.

On ScreenSpot-v2~\citep{screenspot_v2} and the challenging ScreenSpot-Pro~\citep{screenspotpro}, V2P achieves \textbf{92.4\%} and \textbf{52.5\%} element accuracy, significantly outperforming previous methods (see Fig.~\ref{fig:main_results_charts}). Ablation studies confirm that both components consistently contribute to performance gains, demonstrating V2P's broad applicability to high-precision GUI grounding.

Our contribution can be summarized as follows:
\begin{enumerate}[leftmargin=*, label=\arabic*.]
    \item We systematically analyze existing attention-based methods for visual grounding in GUI agents and, through statistical evaluation, identify two main issues——\textit{Background Distraction} and \textit{Center-Edge Confusion}. In addition, we provide a detailed analysis of the underlying causes of these issues and provide insights for further improvements.
    \item We introduce \textit{Attention Suppression Mechanism (SA)} to mitigate Background Distraction and employ \textit{Fitts-Gaussian Peak Modeling (FGPM)} to effectively alleviate Center-Edge Confusion. Building on these methods, we propose the \textbf{Valley-to-Peak (V2P)} framework, an agentic learning paradigm for GUI grounding that significantly enhances the localization precision and accuracy of Vision-Language Models on GUI elements.
    \item Extensive experiments demonstrate that V2P achieves advanced performance on multiple public benchmarks, reaching 92.4\% on ScreenSpot-v2 and 52.50\% on the challenging ScreenSpot-Pro, with relative improvements of 3.6\% and 25.7\%. Furthermore, we confirm that V2P demonstrates significant practical value for real-world deployment and seamless integration into GUI agents.
\end{enumerate}
\section{Related Work}

\subsection{GUI-Agents}
GUI agents have progressed from rudimentary random- or rule-based test tools to multimodal, LLM-driven systems that can follow natural-language instructions. Early efforts such as Monkey testing~\citep{7515500} and planning or script record-and-replay frameworks~\citep{908959, 143213} provided basic coverage but required hand-crafted rules or scripts. Machine-learning techniques later enabled more adaptive behaviour: Humanoid~\citep{li2020humanoiddeeplearningbasedapproach} and Deep GUI~\citep{SE51524} learned user-like action policies from screenshots, while widget detectors~\citep{3293882} improved element recognition. Natural-language interfaces soon followed, e.g. FLIN~\citep{mazumder2021flinflexiblenaturallanguage}  and RUSS~\citep{xu2021groundingopendomaininstructionsautomate}, and reinforcement learning environments like WoB~\citep{pmlr-v70-shi17a} and WebShop~\citep{yao2023webshopscalablerealworldweb} pushed web-scale interaction. The recent arrival of LLMs has unified perception, reasoning and control: WebAgent~\citep{gur2024realworldwebagentplanninglong} and WebGUM~\citep{furuta2024multimodalwebnavigationinstructionfinetuned} achieve open-world browsing, AutoDroid~\citep{wen2024autodroidllmpoweredtaskautomation} and AppAgent~\citep{zhang2023appagentmultimodalagentssmartphone} automate smartphones, and desktop agents such as UFO~\citep{zhang2024ufouifocusedagentwindows} demonstrate GPT-4-level capabilities; industrial systems (e.g. Claude 3.5 Sonnet and Operator) further attest to the practical traction of GUI agents.

\subsection{GUI Grounding}
Prevalent approaches in GUI grounding typically frame the problem as a coordinate generation task~\citep{gui_survey}. Models such as UI-TARS~\citep{ui-tars} and CogAgent~\citep{cogagent} utilize massive supervised fine-tuning to train VLMs to autoregressively generate textual numerical coordinates to ground the target element. However, treating spatial coordinates as ordinary language tokens can limit fine-grained visual alignment. Consequently, recent methods have largely shifted to leveraging the cross-modal attention maps of Vision-Language Models (VLMs)~\citep{gui-actor}. In this paradigm, the model's prediction is derived from the image patch with the highest attention score in response to a language command. While more robust, this approach often suffers from imprecise attention, with focus leaking into irrelevant background regions or spreading too uniformly across the target element. Our work directly addresses this by refining the quality of the attention map itself.

Our approach, V2P, draws inspiration from two distinct areas. To create attention "valleys" and suppress background noise, we adopt attention suppression techniques that penalize focus outside the target region~\citep{suppression-attention}. To form a sharp "peak" at the target's center, we are inspired by both Fitts' Law from Human-Computer Interaction (HCI)~\citep{gauss_1} and the common practice of using Gaussian heatmaps in localization tasks like pose estimation~\citep{gauss_2}. To our knowledge, our work is the first to synergistically combine background suppression with center-focused peak modeling to simulate the human pattern of interaction with the UI elements.
\section{Method}
\label{sec:method}

We introduce Valley-to-Peak (V2P), a method that reshapes the model's attention landscape to mimic human focus patterns for precise GUI grounding. It achieves this through two synergistic components:

\begin{itemize}
    \item \textbf{Suppression Attention Valley Constraint:} Penalizes attention on irrelevant regions to form low-attention "valleys," effectively suppressing background distractions.
    
    \item \textbf{Fitts-Gaussian Peak Modeling:} Models interaction likelihood with a size-adaptive 2D Gaussian, creating a sharp attention "peak" at the target's most actionable center.
\end{itemize}

By jointly optimizing these objectives, V2P produces a continuous, spatially-aware attention map that overcomes the limitations of rigid, uniform labels used in prior work. Below, we first outline the overall architecture (Sec.~\ref{sec:arch}), then detail the Suppression Attention (Sec.~\ref{sec:sup_attn}) and Fitts-Gaussian Peak Modeling (Sec.~\ref{sec:gaussian_peak_modeling}) components.

\begin{figure*}[t]
    \centering
    \includegraphics[width=0.9\linewidth]{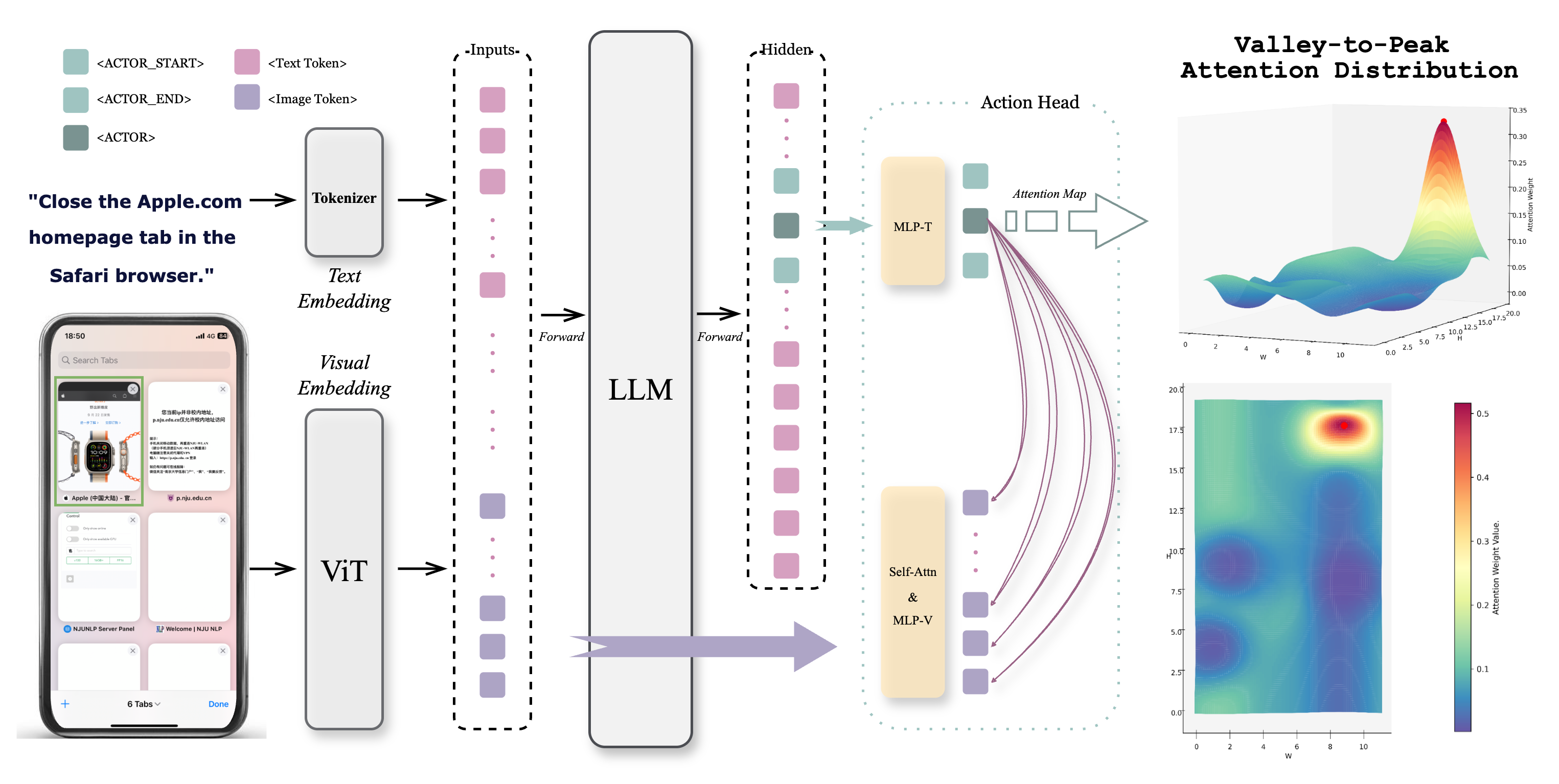}
    \caption{\textbf{Valley-to-Peak training method (V2P).} V2P jointly suppresses noise and enhances signals via two strategies: An inverse-attention penalty carves valleys in non-target areas, while size-adaptive Fitts-Gaussian peaks create sharp peaks at UI elements' centers. This dual approach reshapes attention maps (rightmost example), enabling the model to quickly pinpoint interaction points in cluttered interfaces.}
    \label{fig:main}
\end{figure*}

\subsection{Model Architecture Overview}
\label{sec:arch}

We build upon GUI-Actor~\citep{gui-actor}, a coordinate-free visual grounding framework that localizes GUI actions through attention rather than coordinate regression. Given a screenshot $I$ and an instruction $q$, the model introduces a special token \texttt{<ACTOR>} in the output sequence as a contextual anchor. The final-layer hidden state of \texttt{<ACTOR>}, denoted $h_{\mathtt{<ACTOR>}}$, is used to compute action attention over image patch features $\{v_1, \dots, v_M\}$ extracted by the vision encoder.

To enhance spatial coherence among visual patches, we apply a self-attention module over the patch features:
\begin{equation}
    \tilde{v}_1, \dots, \tilde{v}_M = \text{SelfAttn}(v_1, \dots, v_M)
\end{equation}
yielding contextualized representations. These are projected into a shared embedding space with $h_{\mathtt{<ACTOR>}}$ via separate MLPs:
\begin{align}
    z &= \text{MLP}_T(h_{\mathtt{<ACTOR>}}), \\
    z_i &= \text{MLP}_V(\tilde{v}_i), \quad i = 1,\dots,M.
\end{align}
Attention scores are then computed as:
\begin{equation}
\begin{split}
    \alpha_i &= \frac{z^\top z_i}{\sqrt{d}}\\
    a_i &= \frac{\exp(\alpha_i)}{\sum_{j=1}^M \exp(\alpha_j)}
\end{split}
\end{equation}
where $d$ is the embedding dimension. The resulting $\{a_i\}_{i=1}^M$ forms a normalized attention distribution over the $M$ image patches, representing the model's belief about the target interaction location.

\subsection{Suppression Attention Constraint for Distraction Mitigation}
\label{sec:sup_attn}

Attention maps in complex interfaces can suffer from \emph{attention leakage}, where notable responses are mistakenly assigned to regions far from the target area, particularly in the presence of visually similar distracting patches. To address this issue and enhance spatial precision, we propose a Suppression Attention Constraint. This mechanism explicitly penalizes attention allocated to non-target regions, enforcing sparsity and improving the model's ability to distinguish targets from surrounding distractions.

Let $\mathcal{G} \subset \{1, \dots, M\}$ denote the set of patch indices whose spatial support $R_i$ has empty intersection with the ground-truth bounding box $b$:
\begin{equation}
    \mathcal{G} = \left\{ i \in \{1, \dots, M\} \mid R_i \cap b = \emptyset \right\}
    \label{eq:background_set}
\end{equation}
We define the attention loss as the total attention mass over these irrelevant regions:
\begin{equation}
    \mathcal{L}_{\text{Attn}} = \sum_{i \in \mathcal{G}} a_i
    \label{eq:attn_negative}
\end{equation}

To better understand the theoretical foundation of this constraint, we analyze the gradient dynamics of attention weights. For the target patch $k$ with attention weight $A_k = \text{softmax}(s_k)$, the gradient with respect to any non-target patch logit $s_i$ is:
\begin{equation}
\begin{split}
    w_i &= \frac{\partial A_k}{\partial s_i} = \frac{\partial \text{softmax}(s_k)}{\partial s_i} \\
    &= -\frac{e^{s_k}e^{s_i}}{(\sum_i^M e^{s_i})^2} = -A_k A_i < 0 \quad (i \neq k).
\end{split}
\end{equation}

This gradient analysis reveals that any increase in attention logits $s_i$ for non-target patches negatively impacts the target attention $A_k$. The magnitude $|w_i| = A_k A_i$ quantifies this negative influence: larger values indicate that even small increases in attention to patch $i$ will cause rapid degradation in target attention $A_k$. This theoretical insight naturally motivates using $|w_i|$ as a weighting factor in our suppression loss, providing stronger penalties for patches that pose greater threats to target attention focus. And we have the \textit{suppression attention loss} combined with gradient weight as:
\begin{equation}
    \mathcal{L}_{\text{Sup\_Attn}} = \sum_{i \in \mathcal{G}} w_ia_i
    \label{eq:loss_negative}
\end{equation}

This loss encourages the model to suppress attention on irrelevant regions, thereby reducing the impact of distracting elements in cluttered interfaces. By explicitly minimizing $\mathcal{L}_{\text{Sup\_Attn}}$, the model is incentivized to concentrate its focus on the target region, resulting in enhanced spatial precision and improved robustness.

\subsection{Fitts-Gaussian Peak Modeling for Center-Focused Grounding}
\label{sec:gaussian_peak_modeling}

While the Suppression Attention Constraint encourages focus on target regions, overlapping UI elements can still lead to attention dispersion—particularly toward the boundaries of positively labeled components—resulting in ambiguous and spatially diffused attention maps.

Our supervision strategy is inspired by Fitts' Law~\citep{gauss_1, gauss_2}, which reveals that click probability peaks at the center of an UI element and decays toward its edges, closely following a Gaussian distribution. We encode this behavior with Fitts-Gaussian Peak Modeling to guide the model's focus in line with observed human interaction.

Specifically, we model the ideal attention distribution as a 2D Gaussian density centered at the centroid of the ground-truth bounding box $b = [x_1, y_1, x_2, y_2]$:
\begin{equation}
    \mu = (c_x, c_y) = \left( \frac{x_1 + x_2}{2}, \frac{y_1 + y_2}{2} \right)
\end{equation}

To reflect the interaction tolerance associated with target size, we set the standard deviation of the Gaussian proportional to the element’s width and height:
\begin{equation}
    \sigma_x = \frac{w}{\sigma_{\text{factor}}}, \quad
    \sigma_y = \frac{h}{\sigma_{\text{factor}}}
\end{equation}
where $w = x_2 - x_1$, $h = y_2 - y_1$, and $\sigma_{\text{factor}}$ is a hyperparameter controlling the concentration of the attention prior. This formulation ensures that larger elements—more tolerant to pointing errors—induce broader attention peaks, while smaller elements require sharper focus.

Given an input image partitioned into $M = H \times W$ non-overlapping patches of size $s \times s$, we compute the expected attention mass for each patch $i$, covering spatial region $R_i = [x^i_{\min}, x^i_{\max}] \times [y^i_{\min}, y^i_{\max}]$, by integrating the 2D Gaussian density over $R_i$:
\begin{equation}
    y_i = \int_{R_i} \mathcal{N}(x, y; \mu, \Sigma)  dx\,dy
\end{equation}
where $\Sigma = \mathrm{diag}(\sigma_x^2, \sigma_y^2)$. Thanks to axis-aligned separability, this integral decomposes efficiently into the product of two univariate cumulative distribution functions (CDFs):
\begin{equation}
\begin{split}
    y_i &= \left[ \Phi(x^i_{\max}; c_x, \sigma_x) - \Phi(x^i_{\min}; c_x, \sigma_x) \right] \\
    &\quad \cdot \left[ \Phi(y^i_{\max}; c_y, \sigma_y) - \Phi(y^i_{\min}; c_y, \sigma_y) \right]
\end{split}
\end{equation}
with $\Phi(\cdot\,; \mu, \sigma)$ denoting the CDF of a univariate normal distribution.

To supervise the model's predicted attention distribution $\{a_i\}$, we adopt the action attention loss from GUI-Actor~\citep{{gui-actor}}, using the Kullback-Leibler (KL) divergence to measure the discrepancy between the target $p$ and prediction $a$:

\begin{equation}
\begin{split}
    \mathcal{L}_{\text{Action\_Attn}} &= \sum_{i=1}^M p_i \log \frac{p_i}{a_i}, \\
    p_i &= \frac{y_i}{\sum_{j=1}^M y_j + \epsilon},\\
    &\quad i = 1, \ldots, M
\end{split}
    \label{eq:loss_attn}
\end{equation}

where $\epsilon$ is a small constant for numerical stability.

Fitts-Gaussian Peak Modeling establishes a center-biased, size-aware attention prior that closely mimics human pointing behavior. By discouraging boundary leakage and promoting centralized attention in a graded, interaction-informed manner, it enhances localization precision and improves robustness in complex and cluttered UI layouts.

\subsection{Valley-to-Peak Training}
The overall training objective combines next-token prediction loss with action-focused attention losses:
\begin{equation}
    \mathcal{L} = \mathcal{L}_{\text{NTP}} + \lambda_1 \mathcal{L}_{\text{Sup\_Attn}} + \lambda_2 \mathcal{L}_{\text{Action\_Attn}}
    \label{eq:total_loss}
\end{equation}

where $\mathcal{L}_{\text{Sup\_Attn}}$ suppresses attention outside the target region (Section~\ref{sec:sup_attn}), and $\mathcal{L}_{\text{Action\_Attn}}$ enforces alignment between predicted attention and a Gaussian-shaped target distribution (Section~\ref{sec:gaussian_peak_modeling}).

Minimizing the combined loss supports a \emph{Valley-to-Peak} training paradigm: coarse suppression followed by fine-grained alignment. $\mathcal{L}_{\text{Sup\_Attn}}$ first suppresses distractions, guiding attention toward the target region. Then, $\mathcal{L}_{\text{Action\_Attn}}$ sharpens this focus by prioritizing the target’s center. This reduces misclicks and alleviates ambiguity caused by overlapping labels, ensuring precise and human-like attention alignment. The coarse-to-fine control enables robust interaction predictions, even in dense and visually complex UI environments.

\section{Experiment}

\subsection{Experimental Setup}
\label{sec:setup}
We utilize Qwen2.5-VL-Instruct (both 7B and 3B)~\citep{Qwen2.5-VL} as backbones. To ensure a rigorously fair comparison and isolate algorithmic contributions, we strictly follow the data recipe of the baseline GUI-Actor~\citep{gui-actor}, with a learning rate of 5e-6 and $\sigma=1.0$. Comprehensive details are provided in App.~\ref{app:train_details}.

We evaluate on a comprehensive suite of six benchmarks. Our primary evaluation focuses on \textit{ScreenSpot-v2}~\citep{screenspot_v2} and \textit{ScreenSpot-Pro}~\citep{screenspotpro}, as they provide the most standardized assessment across diverse platforms and challenging high-resolution OOD scenarios. 

To further verify robustness and agentic potential, we also test on \textit{OSWorld-G}~\citep{OSWorld}, \textit{UI-Vision} (Element Grounding)~\citep{UI-Vision}, \textit{UI-I2E}~\citep{UI-I2E}, and \textit{MMBench-GUI L2}~\citep{MMBench}.

\subsection{Main Results}
\label{sec:main_results}
Tab.~\ref{tab:main_results} presents a comprehensive evaluation of V2P against other baselines.

\textbf{Superior Performance on ScreenSpot Benchmarks.} 
As our primary evaluation field, V2P-7B demonstrates exceptional capabilities among models of similar scale.
On \textit{ScreenSpot-v2}, it achieves a competitive accuracy of 92.4\%. 
More critically, on the high-difficulty \textit{ScreenSpot-Pro}, which features high-resolution screens and OOD applications, V2P-7B attains 52.5\%, significantly outperforming the strong baseline GUI-Actor-7B (44.6\%) and UI-TARS-72B (38.1\%). 
This substantial margin validates that V2P's attention calibration is particularly effective in handling the dense, visually complex interfaces typical of professional GUI environments.

\textbf{Generalization to Agentic Scenarios.} 
To assess the model's potential as a perception backend for autonomous agents, we extend our evaluation to four benchmarks featuring interaction traces and functional reasoning requirements: \textit{OSWorld-G}~\citep{OSWorld}, \textit{UI-Vision} (Element Grounding)~\citep{UI-Vision}, \textit{UI-I2E}~\citep{UI-I2E}, and \textit{MMBench-GUI L2}~\citep{MMBench}.
As shown in Tab.~\ref{tab:main_results}, V2P-7B demonstrates superior performance across the majority of evaluations. 
Notably, V2P-7B surpasses all other baselines on \textit{UI-Vision}, \textit{UI-I2E}, and \textit{MMBench-GUI L2}. This consistent superiority highlights the model's exceptional functional reasoning and semantic understanding.
Furthermore, on \textit{OSWorld-G}, V2P matches the specialist JEDI-7B~\citep{OSWorld} (52.5\%) despite using only $\sim$50k PC samples versus JEDI's millions. Moreover, V2P significantly surpasses JEDI on other benchmarks, highlighting superior data efficiency and generalization beyond specific domains.

\textbf{Scalability and Efficiency.}
As shown in the \textit{Controlled Comparison} group, V2P-3B consistently outperforms its direct competitor GUI-Actor-3B across all six benchmarks. Notably, on some challenging benchmarks, it even surpasses significantly larger scale models. This result underscores the pure algorithmic superiority of the V2P framework and its consistent effectiveness across varying model scales.

\begin{table*}[t]
    \centering
    \small
    \setlength{\tabcolsep}{3pt}
    \renewcommand{\arraystretch}{1.0}
    \setlength{\abovecaptionskip}{2pt}
    \setlength{\belowcaptionskip}{-4pt}
    \resizebox{\textwidth}{!}{
    \begin{tabular}{@{}l@{\hspace{2pt}}|@{\hspace{4pt}}r@{\hspace{4pt}}r@{\hspace{4pt}}|@{\hspace{4pt}}r@{\hspace{4pt}}r@{\hspace{4pt}}r@{\hspace{4pt}}r@{}}
    \toprule
    \multirow{2}{*}{\textbf{Model}} & \multicolumn{2}{c|}{\textbf{General Grounding}} & \multicolumn{4}{c}{\textbf{Complex \& Semantic Grounding}} \\
    \cmidrule(lr){2-3} \cmidrule(lr){4-7}
    & \textbf{ScreenSpot-v2} & \textbf{ScreenSpot-Pro} & \textbf{OSWorld-G} & \textbf{UI-Vision} & \textbf{UI-I2E} & \textbf{MMBench} \\
    \midrule
    \multicolumn{7}{c}{\textit{Proprietary \& General VLMs}} \\
    \addlinespace[0.1em]
    GPT-4o & 80.7 & 0.8 & -- & 1.38 & -- & 2.87 \\
    Operator & 70.5 & -- & 40.6 & -- & -- & -- \\
    Qwen2.5-VL-3B & 80.9 & 16.1 & 27.3 & -- & 41.7 & -- \\
    Qwen2.5-VL-7B & 88.8 & 26.8 & 31.4 & 0.85 & 53.8 & 33.9 \\
    \midrule
    \multicolumn{7}{c}{\textit{GUI-Specialized Models (SFT)}} \\
    \addlinespace[0.1em]
    SeeClick-9.6B & 55.1 & 1.1 & -- & 5.39 & 26.4 & -- \\
    OS-Atlas-7B & 84.1 & 18.9 & 27.7 & 9.02 & 58.6 & 41.4 \\
    Aguvis-7B & 86.0 & 22.9 & 38.7 & 13.7 & 53.2 & 45.7 \\
    UGround-V1-7B & 87.6 & 31.1 & 36.4 & 12.9 & 70.3 & 65.7 \\
    UI-TARS-7B & 91.6 & 35.7 & 47.5 & 17.6 & 61.4 & 64.3 \\
    JEDI-7B & 91.7 & 39.5 & 54.1 & 24.8 & -- & -- \\
    UI-TARS-72B & 90.3 & 38.1 & \textbf{57.1} & 25.5 & 73.7 & 74.3 \\
    \midrule
    \multicolumn{7}{c}{\textit{Controlled Comparison (Identical Training Data)}} \\
    \addlinespace[0.1em]
    GUI-Actor-3B & 91.0 & 42.2 & 45.9 & 21.9 & 63.7 & 73.5 \\
    \rowcolor{gray!10} \textbf{V2P-3B (Ours)} & \textbf{91.4} & \textbf{48.5} & \textbf{48.8} & \textbf{26.0} & \textbf{69.5} & \textbf{77.6} \\
    \cmidrule(lr){1-7}
    GUI-Actor-7B & 92.1 & 44.6 & 49.3 & 24.3 & 68.2 & 76.5 \\
    \rowcolor{gray!10} \textbf{V2P-7B (Ours)} & \textbf{92.4} & \textbf{52.5} & \textbf{52.5} & \textbf{28.8} & \textbf{75.6} & \textbf{79.9} \\
    \bottomrule
    \end{tabular}
    }
    \caption{\textbf{Main Results Comparison.} We evaluate V2P against state-of-the-art baselines across six diverse benchmarks, covering general, high-resolution, and agentic GUI scenarios. V2P-7B significantly outperforming baselines under comparable settings.}
    \label{tab:main_results}
\end{table*}

\subsection{Ablation and Analysis}
\label{sec:analysis}
\subsubsection{Component Ablation Study}
To validate the necessity of our proposed modules, we conducted a standard ablation study on \textit{ScreenSpot-Pro} (Tab.~\ref{tab:ablation_component}). Removing \textit{Fitts-Gaussian Peak Modeling (FGPM)} leads to a significant performance drop of 5.0\%, confirming its critical role in precise localization. Further removing \textit{Suppression Attention (SA)} results in an additional loss of 3.2\%. These results verify that both modules are indispensable for the V2P framework.

\begin{table}[h]
\centering
\small
\setlength{\tabcolsep}{4pt}
\setlength{\abovecaptionskip}{2pt}
\setlength{\belowcaptionskip}{-4pt}
\renewcommand{\arraystretch}{1.0}
\begin{tabular}{l|cc}
\toprule
\textbf{Model Variant} & \textbf{Pro Avg.} & \textbf{$\Delta$} \\
\midrule
\textbf{V2P-7B (Full)} & \textbf{52.5} & - \\
\quad w/o FGPM & 47.5 & \textcolor{red}{-5.0} \\
\quad w/o FGPM \& SA & 44.3 & \textcolor{red}{-8.2} \\
\bottomrule
\end{tabular}
\caption{\textbf{Component Ablation on ScreenSpot-Pro.} Both FGPM and SA contribute significantly to the final performance.}
\label{tab:ablation_component}
\end{table}

\subsubsection{Attribution of Performance Gains}
\label{sec:attribution_analysis}
To investigate the underlying reasons for V2P's superior performance, we conducted a quantitative performance gains attribution analysis on 182 samples where V2P-7B successfully corrected the failures of the baseline GUI-Actor-7B~\citep{gui-actor}. 
As shown in Tab.~\ref{tab:attribution}, the results reveal that 50.5\% of the performance gains stem from effectively suppressing \textit{Background Distraction}, while 35.7\% are attributed to resolving \textit{Center-Edge Confusion}. This provides strong empirical evidence that V2P's dual-loss mechanism functions exactly as designed.

\begin{table}[h]
\centering
\small
\setlength{\tabcolsep}{4pt}
\setlength{\abovecaptionskip}{2pt}
\setlength{\belowcaptionskip}{-4pt}
\renewcommand{\arraystretch}{1.0}
\resizebox{\columnwidth}{!}{
\begin{tabular}{l|c|c}
\toprule
\textbf{Baseline Error Type} & \textbf{Count} & \textbf{Contribution} \\
\midrule
Background Distraction & 92 & \textbf{50.5\%} \\
Center-Edge Confusion & 65 & \textbf{35.7\%} \\
Other / Normal Attention & 25 & 13.7\% \\
\midrule
Total Improved Samples & 182 & 100\% \\
\bottomrule
\end{tabular}
}
\caption{\textbf{Performance Gains Attribution Analysis.} We analyzed samples from ScreenSpot-Pro where V2P-7B made correct predictions while the baseline GUI-Actor~\citep{gui-actor} failed. The majority of gains come from correcting background and center-edge errors.}
\label{tab:attribution}
\end{table}

\begin{figure*}[!htbp]
    \centering
    \vspace{-0.2cm}
    \includegraphics[width=0.88\textwidth]{img/gauss_factor_and_generalization_original.png}
    \vspace{-0.3cm}
    \caption{\textbf{Gaussian Factor and Generalization Ability Analysis.} (a) Impact of Gaussian Factor $\sigma$. A smaller $\sigma$ (sharper peak) benefits precision, with the optimal performance achieved at $\sigma=1.0$ for ScreenSpot-Pro. Larger $\sigma$ values degrade performance due to introduced label noise. (b, c) Generalization Ability. V2P shows consistent improvement, whereas the baseline suffers from overfitting on OOD data.}
    \label{fig:ablation_sigma}
    \end{figure*}

\subsubsection{Performance Leap on Tiny Targets}
\label{sec:size_analysis}
To evaluate performance on fine-grained targets, we categorized UI elements across \textit{ScreenSpot-v2} and \textit{ScreenSpot-Pro} based on their area relative to the patch size $n$ ($14 \times 14$). Specifically, elements are classified as Small ($n \le A < 4n$), Medium ($4n \le A < 9n$), and Large ($A \ge 9n$). As shown in Tab.~\ref{tab:size_stratified}, V2P-7B outperforms the baseline GUI-Actor-7B~\citep{gui-actor} by 10.0\% on these small elements. This demonstrates the superiority of V2P in the fine-grained positioning of small targets.

Furthermore, the data shown in Tab.~\ref{tab:size_stratified} also reveals a critical distribution shift between benchmarks: \textit{ScreenSpot-v2} is dominated by large elements (size $>9n$), which offer vast spatial tolerance. Consequently, even spatially diffuse attention maps often fall within these generous boundaries, which explains the high accuracy of the baseline on \textit{ScreenSpot-v2}, effectively masking its inherent localization imprecision. In contrast, \textit{ScreenSpot-Pro} is densely populated with small elements that tolerate negligible error. Consequently, V2P-7B's precision advantage, while masked on the coarse-grained \textit{ScreenSpot-v2}, is fully realized on the challenging \textit{ScreenSpot-Pro}.

\subsubsection{Sensitivity to Gaussian Factor $\sigma$}
\label{sec:sigma_analysis}

To analyze the impact of the Gaussian factor $\sigma$ on grounding precision, we conducted ablation experiments on \textit{ScreenSpot-v2} and \textit{ScreenSpot-Pro} across varying $\sigma$ values. As shown in Fig.~\ref{fig:ablation_sigma}(a), model performance is strongly sensitive to this hyperparameter. On \textit{ScreenSpot-v2}, accuracy improves from 91.3\% ($\sigma=6.0$) to 92.4\% ($\sigma=0.5$). Similarly, \textit{ScreenSpot-Pro} achieves its peak accuracy of 52.5\% at $\sigma=1.0$, while larger $\sigma$ values cause a significant decline.

\begin{table}[h]
\centering
\small
\setlength{\tabcolsep}{4pt}
\setlength{\abovecaptionskip}{2pt}
\setlength{\belowcaptionskip}{-4pt}
\renewcommand{\arraystretch}{1.0}
\resizebox{\columnwidth}{!}{
\begin{tabular}{l|cc|cc}
\toprule
\multirow{2}{*}{\textbf{Element Size}} & \multicolumn{2}{c|}{\textbf{ScreenSpot-v2}} & \multicolumn{2}{c}{\textbf{ScreenSpot-Pro}} \\
 & \textit{GUI-Actor} & \textit{V2P} & \textit{GUI-Actor} & \textbf{V2P} \\
\midrule
\textbf{Small} ($n \sim 4n$) & 50.0\% & \textbf{60.0\%} & 17.5\% & \textbf{23.8\%} \\
\textbf{Medium} ($4n \sim 9n$) & 71.4\% & \textbf{85.7\%} & 43.1\% & \textbf{47.9\%} \\
\textbf{Large} ($> 9n$) & 93.2\% & 92.9\% & 60.3\% & \textbf{66.6\%} \\
\bottomrule
\end{tabular}
}
\caption{\textbf{Size-stratified Performance.} V2P achieves substantial gains on small elements in ScreenSpot-v2 and ScreenSpot-Pro, underscoring its superior capability in precise fine-grained localization.}
\label{tab:size_stratified}
\end{table}

We attribute this phenomenon to the spatial concentration of the attention mechanism. Larger $\sigma$ values generate broader Gaussian distributions, which tend to dilute the spatial focus and introduce background noise into the attention maps. Conversely, a smaller $\sigma$ produces sharper Gaussian peaks. This acts as a tight spatial constraint, allowing the model to localize UI elements with higher precision and resulting in more accurate click predictions. These results underscore the necessity of balancing $\sigma$: while excessively large values hinder localization, a moderately small $\sigma$ (e.g., 1.0) significantly enhances spatial accuracy.

\subsubsection{Training Stability and Generalization}
\label{sec:stability_analysis}
Finally, we evaluate the training stability of V2P-7B compared to the Aguvis-7B~\citep{Aguvis}. 
As visualized in Fig.~\ref{fig:ablation_sigma}(b) and (c), V2P-7B demonstrates a consistently ascending accuracy curve on both in-distribution (\textit{ScreenSpot-v2}) and out-of-distribution (\textit{ScreenSpot-Pro}) benchmarks.
In sharp contrast, Aguvis-7B~\citep{Aguvis} exhibits a distinct "overfitting-to-distribution" pattern: while its performance improves on \textit{ScreenSpot-v2}, it suffers from a continuous performance decline on the OOD \textit{ScreenSpot-Pro} after the 20\% training milestone. 
This confirms that our human-like visual attention mechanism (\textit{Fitts-Gaussian Peak Modeling} and \textit{Suppression Attention}) effectively mitigates the overfitting inherent to textual coordinate supervision, ensuring robust generalization across unseen scenarios.

\section{Conclusion}

In this paper, we address the critical bottlenecks of \textit{Background Distraction} and \textit{Center-Edge Confusion} in GUI grounding by proposing \textbf{Valley-to-Peak (V2P)} framework. Mimicking human visual processing, V2P synergizes \textit{Suppression Attention} to eliminate background noise and \textit{Fitts-Gaussian Peak Modeling} to construct sharp, size-adaptive peaks at actionable centers.

By emulating this human-like strategy for visual localization, our approach fosters a more authentic spatial understanding of complex interfaces. Extensive experiments confirm the effectiveness of this framework: V2P achieves exceptional results on ScreenSpot-v2 (92.4\%) and the challenging ScreenSpot-Pro (52.5\%), consistently outperforming existing strong baselines. Notably, our method demonstrates remarkable robustness on fine-grained small targets and out-of-distribution scenarios, effectively bridging the gap between coarse perception and precise actuation. By enabling agents to "see" and "focus" like human users, V2P offers a scalable and robust foundation for the next generation of general-purpose GUI agents.
\section*{Limitations}

While V2P demonstrates exceptional performance across various benchmarks, several limitations remain to be addressed:

\begin{itemize}
    \item \textbf{Ambiguity among Semantically Similar Targets:} As analyzed in our failure case studies (see App.~\ref{app:qualitative_analysis}), the model occasionally struggles when multiple UI elements share high semantic similarity, such as identical icons with different functional purposes. This suggests that visual calibration alone may not fully resolve deep logical intent without more comprehensive UI context.
    
    \item \textbf{Generalization to Unconventional Designs:} The model's attention distribution can become highly dispersed when encountering unconventional or cluttered layouts that deviate from the training distribution, indicating uncertainty in complex visual environments.
    
    \item \textbf{Computational Overhead:} The introduction of the self-attention module to enhance spatial coherence among visual patches may introduce marginal increases in inference latency compared to simple coordinate regression methods, particularly when processing high-resolution screenshots with a large number of patches.
\end{itemize}

\section*{Ethics Statement}

In this work, we propose the Valley-to-Peak (V2P) framework to improve GUI grounding by mimicking human visual processing. We adhere to the ACL Code of Ethics and highlight the following:

\begin{itemize}
    \item \textbf{Data Privacy:} All training and evaluation datasets used in this study are from publicly available academic sources. We have strictly followed data recipe guidelines to exclude samples containing personal identifiable information (PII).
    
    \item \textbf{Mitigation of Bias:} Our training data spans multiple operating systems and platforms (Mobile, Desktop, Web) to minimize algorithmic bias toward specific UI design patterns.
\end{itemize}

\section*{Acknowledgements}

The authors would like to thank the anonymous reviewers for their insightful feedback. We acknowledge the use of generative AI tools for polishing the linguistic quality and refining the prose of this manuscript. All technical claims and final content remain the sole responsibility of the authors.

\bibliography{iclr2025_conference}

\appendix
\appendix

\section{Training and Inference Details}
\label{app:train_details}

\subsection{Source Training Data}
Following GUI-Actor~\citep{gui-actor}, we compile our training dataset from several publicly available, high-quality GUI datasets, with summary statistics provided in Tab.~\ref{tab:training_datasets}. To ensure fair evaluation, we also exclude any samples from Wave-UI that overlap with the test sets of downstream tasks. 

\begin{table*}[ht]
    \centering
    \begin{tabular}{lccc}
        \toprule
        \textbf{Dataset} & \textbf{\# of Elements} & \textbf{\# of Screenshots} & \textbf{Platform} \\
        \midrule
        Uground Web–Hybrid~\citep{8}      & 8M    & 775K & Web      \\
        GUI-Env~\citep{23}                    & 262K  & 70K  & Web      \\
        GUI-Act~\citep{23}                    & 42K   & 13K  & Web      \\
        AndroidControl~\citep{55}             & 47K   & 47K  & Android  \\
        AMEX~\citep{24}                       & 1.2M  & 100K & Android  \\
        Wave-UI       & 50K   & 7K   & Hybrid   \\
        \midrule
        \textbf{Total}                       & 9.6M  & 1M   & --       \\
        \bottomrule
    \end{tabular}
    \caption{Overview of training datasets used for GUI-Actor.}
    \label{tab:training_datasets}
\end{table*}

\section{Benchmarks}
\label{app:benchmarks} 
Our evaluation centers on six sophisticated benchmarks for GUI visual grounding:

\textbf{ScreenSpot-v2}~\citep{screenspot_v2} encompasses 1,272 carefully annotated instructions, each paired with corresponding target elements across diverse GUI environments, including mobile (Android and iOS), desktop (macOS and Windows), and web platforms. The dataset is designed to improve the quality and reliability of GUI visual grounding tasks, addressing key challenges such as eliminating ambiguities in natural language instructions and resolving annotation errors. By refining the alignment between textual descriptions and interface elements, ScreenSpot-v2 provides a robust and standardized benchmark for evaluating grounding models.

\textbf{ScreenSpot-Pro}~\citep{screenspotpro}, meanwhile, focuses on more demanding scenarios, especially those involving high-resolution professional applications. It contains 1,581 tasks annotated by domain experts across 23 specialized software applications, spanning three operating systems. This benchmark significantly broadens the scope of GUI visual grounding by introducing interfaces with industrial software and multi-window layouts, creating a larger domain gap compared to most pretraining data. With its increased complexity and domain diversity, ScreenSpot-Pro is an invaluable resource for assessing the generalization ability of models in realistic and challenging GUI environments.

\textbf{OSWorld-G} is the grounding-specific subset derived from the OSWorld benchmark~\citep{OSWorld}, a unified evaluation environment for multimodal agents on Ubuntu. Unlike static datasets, OSWorld-G consists of screenshots captured from a fully functional, interactive operating system. It evaluates the model's ability to localize actionable elements within dynamic and complex real-world desktop workflows, serving as a direct proxy for an agent's practical utility in autonomous computer control tasks.

\textbf{UI-Vision (Element Grounding)}~\citep{UI-Vision} is designed to rigorously test the semantic understanding of user interface elements. While standard grounding tasks often rely on text matching (OCR), UI-Vision focuses on functional icons and visual symbols (e.g., identifying a "magnifying glass" as "search" or a "floppy disk" as "save") that lack explicit textual labels. Performance on this benchmark reflects the model's capacity for visual reasoning and its ability to interpret the functional affordances of GUI components.

\textbf{UI-I2E} (Image-to-Element)~\citep{UI-I2E} evaluates the capability to parse the hierarchical structure of a screen. The task requires the model to map raw pixel inputs to structured representations, effectively "reading" the underlying layout or accessibility tree of the interface. High accuracy on UI-I2E indicates that the model possesses a deep understanding of UI composition and element spatial relationships, rather than merely memorizing surface-level patterns.

\textbf{MMBench-GUI L2}~\citep{MMBench} is the GUI-specific subset (L-2 category) of the massive MMBench suite. Adopting a robust CircularEval strategy with multiple-choice questions, it assesses fine-grained perception and reasoning abilities within graphical interfaces. This benchmark serves as a standardized indicator of the model's general-purpose multimodal intelligence in the GUI domain, complementing the pure localization metrics of ScreenSpot.

\begin{table*}[!t]
    \centering
    \small
    \setlength{\tabcolsep}{4pt}
    \setlength{\abovecaptionskip}{3pt}
    \setlength{\belowcaptionskip}{-8pt}
    \renewcommand{\arraystretch}{0.9}
    \setlength{\extrarowheight}{0pt}
    
        \begin{tabular}{@{}l*{7}{c}@{}}
        \toprule
        \multirow{3}{*}{Model} & \multicolumn{7}{c}{ScreenSpot-v2 Accuracy (\%)} \\
        \cmidrule(lr){2-8}
        \cmidrule(lr){2-8} & \multicolumn{1}{c}{Mobile-Text} & \multicolumn{1}{c}{Mobile-Icon} & \multicolumn{1}{c}{Desktop-Text} & \multicolumn{1}{c}{Desktop-Icon} & \multicolumn{1}{c}{Web-Text} & \multicolumn{1}{c}{Web-Icon} & \multicolumn{1}{c}{Avg.} \\
        \midrule
        \multicolumn{8}{c}{\textbf{\textit{Proprietary Models}}}\\
        \addlinespace
        Operator                                & 47.3 & 41.5 & 90.2 & 80.3 & 92.8 & 84.3 & 70.5 \\
        GPT-4o + OmniParser-v2                  & 95.5 & 74.6 & 92.3 & 60.9 & 88.0 & 59.6 & 80.7 \\
        \midrule
        \multicolumn{8}{c}{\textbf{\textit{General Open-source Models}}}\\
        \addlinespace
        Qwen2.5-VL-3B                           & 93.4 & 73.5 & 88.1 & 58.6 & 88.0 & 71.4 & 80.9 \\
        Qwen2.5-VL-7B                           & 97.6 & 87.2 & 90.2 & 74.2 & 93.2 & 81.3 & 88.8 \\
        \midrule
        \multicolumn{8}{c}{\textbf{\textit{GUI-specific Models (SFT)}}}\\
        \addlinespace
        SeeClick-9.6B                           & 78.4 & 50.7 & 70.1 & 29.3 & 55.2 & 32.5 & 55.1 \\
        Magma-8B                                & 62.8 & 53.4 & 80.0 & 57.9 & 67.5 & 47.3 & 61.5 \\
        OS-Atlas-4B                             & 87.2 & 59.7 & 72.7 & 46.4 & 85.9 & 63.1 & 71.9 \\
        UI-TARS-2B                              & 95.2 & 79.1 & 90.7 & 68.6 & 87.2 & 78.3 & 84.7 \\
        OS-Atlas-7B                             & 95.2 & 75.8 & 90.7 & 63.6 & 90.6 & 77.3 & 84.1 \\
        Aguvis-7B                               & 95.5 & 77.3 & 95.4 & 77.9 & 91.0 & 72.4 & 86.0 \\
        UGround-V1-7B                           & 95.0 & 83.3 & 95.0 & 77.8 & 92.1 & 77.2 & 87.6 \\
        UI-TARS-72B                             & 94.8 & 86.3 & 91.2 & 87.9 & 91.5 & 87.7 & 90.3 \\
        GUI-Actor-3B                            & 97.6 & 83.4 & 96.9 & 83.6 & 94.0 & 85.7 & 91.0 \\
        UI-TARS-7B                              & 96.9 & 89.1 & 95.4 & 85.0 & 93.6 & 85.2 & 91.6 \\
        GUI-Actor-7B                            & 97.6 & 88.2 & 96.9 & 85.7 & 93.2 & 86.7 & 92.1 \\
        \midrule
        \multicolumn{8}{c}{\textbf{\textit{GUI-specific Models (RL)}}}\\
        \addlinespace
        SE-GUI-7B                               & - & - & - & - & - & - & 90.3 \\
        LPO-8B                           & - & - & - & - & - & - & 90.5 \\
        \midrule
        \multicolumn{8}{c}{\textbf{\textit{Ours}}}\\
        \addlinespace
        \textbf{V2P-7B}                             & \textbf{98.1} & 88.0 & 96.1 & \textbf{89.7} & \textbf{95.4} & 84.4 & \textbf{92.4} \\
        \bottomrule
        \end{tabular}
    \caption{Comparison of Model Performance Across Task Categories in ScreenSpot-v2. Bold text highlights the best results, while ``–'' represents missing values not reported in the original papers.}
    \label{tab:screenspot_v2}
\end{table*}

\section{Detailed Experimental Results on ScreenSpot-v2 and ScreenSpot-Pro}
\label{app:screenspot_details}

We provide extended experimental results, including fine-grained performance breakdowns and comparisons against a broader set of baselines. Detailed statistics are presented in Tab.~\ref{tab:screenspot_v2} and Tab.~\ref{tab:screenspot_pro}.

\begin{table*}[!t]
    \centering
    \footnotesize
    \setlength{\tabcolsep}{4pt}
    \setlength{\abovecaptionskip}{3pt}
    \setlength{\belowcaptionskip}{-6pt}
    \renewcommand{\arraystretch}{0.95}
    
    \resizebox{\textwidth}{!}{%
        \begin{tabular}{lccccccccccccccc}
        \toprule
        \multirow{3}{*}{Model} & \multicolumn{12}{c}{ScreenSpot-Pro Accuracy (\%)} \\
        \cmidrule(lr){2-16} & \multicolumn{2}{c}{CAD} & \multicolumn{2}{c}{Dev} & \multicolumn{2}{c}{Creative} & \multicolumn{2}{c}{Scientific} & \multicolumn{2}{c}{Office} & \multicolumn{2}{c}{OS} & \multicolumn{3}{c}{Avg.}\\
        \cmidrule(lr){2-3}\cmidrule(lr){4-5}\cmidrule(lr){6-7}\cmidrule(lr){8-9}\cmidrule(lr){10-11}\cmidrule(lr){12-13}\cmidrule(lr){14-16}
        & Text & Icon & Text & Icon & Text & Icon & Text & Icon & Text & Icon & Text & Icon & Text & Icon & \textbf{Avg.} \\[1pt]
        \midrule
        \multicolumn{16}{c}{\textbf{\textit{Proprietary Models}}}\\
        \addlinespace
        GPT-4o              & 2.0 & 0.0 & 1.3 & 0.0 & 1.0 & 0.0 & 2.1 & 0.0 & 1.1 & 0.0 & 0.0 & 0.0 & 1.3 & 0.0 & 0.8\\
        Claude Computer Use & 14.5 & 3.7 & 22.0 & 3.9 & 25.9 & 3.4 & 33.9 & 15.8 & 30.1 & 16.3 & 11.0 & 4.5 & 23.4 & 7.1 & 17.1\\
        \midrule
        \multicolumn{16}{c}{\textbf{\textit{General Open-source Models}}}\\
        \addlinespace
        Qwen2.5-VL-3B      & 9.1 & 7.3 & 22.1 & 1.4 & 26.8 & 2.1 & 38.2 & 7.3 & 33.9 & 15.1 & 10.3 & 1.1 & 23.6 & 3.8 & 16.1\\
        Qwen2.5-VL-7B        & 16.8 & 1.6 & 46.8 & 4.1 & 35.9 & 7.7 & 49.3 & 7.3 & 52.5 & 20.8 & 37.4 & 6.7 & 38.9 & 7.1 & 26.8\\
        \midrule
        \multicolumn{16}{c}{\textbf{\textit{GUI-specific Models (SFT)}}}\\
        \addlinespace
        SeeClick-9.6B        & 2.5 & 0.0 & 0.6 & 0.0 & 1.0 & 0.0 & 3.5 & 0.0 & 1.1 & 0.0 & 2.8 & 0.0 & 1.8 & 0.0 & 1.1\\
        FOCUS-2B             & 7.6 & 3.1 & 22.8 & 1.7 & 23.7 & 1.7 & 25.0 & 7.1 & 23.2 & 7.7 & 17.8 & 2.5 & 19.8 & 3.9 & 13.3\\
        CogAgent-18B         & 7.1 & 3.1 & 14.9 & 0.7 & 9.6 & 0.0 & 22.2 & 1.8 & 13.0 & 0.0 & 5.6 & 0.0 & 12.0 & 0.8 & 7.7\\
        Aria-UI              & 7.6 & 1.6 & 16.2 & 0.0 & 23.7 & 2.1 & 27.1 & 6.4 & 20.3 & 1.9 & 4.7 & 0.0 & 17.1 & 2.0 & 11.3\\
        OS-Atlas-7B          & 12.2 & 4.7 & 33.1 & 1.4 & 28.8 & 2.8 & 37.5 & 7.3 & 33.9 & 5.7 & 27.1 & 4.5 & 28.1 & 4.0 & 18.9\\
        ShowUI-2B            & 2.5 & 0.0 & 16.9 & 1.4 & 9.1 & 0.0 & 13.2 & 7.3 & 15.3 & 7.5 & 10.3 & 2.2 & 10.8 & 2.6 & 7.7\\
        UGround-7B           & 14.2 & 1.6 & 26.6 & 2.1 & 27.3 & 2.8 & 31.9 & 2.7 & 31.6 & 11.3 & 17.8 & 0.0 & 25.0 & 2.8 & 16.5\\
        UGround-V1-7B        & 15.8 & 1.2 & 51.9 & 2.8 & 47.5 & 9.7 & 57.6 & 14.5 & 60.5 & 13.2 & 38.3 & 7.9 & 45.2 & 8.1 & 31.1\\
        UI-TARS-2B           & 17.8 & 4.7 & 47.4 & 4.1 & 42.9 & 6.3 & 56.9 & 17.3 & 50.3 & 17.0 & 21.5 & 5.6 & 39.6 & 8.4 & 27.7\\
        UI-TARS-7B           & 20.8 & 9.4 & 58.4 & 12.4 & 50.0 & 9.1 & 63.9 & 31.8 & 63.3 & 20.8 & 30.8 & 16.9 & 47.8 & 16.2 & 35.7\\
        UI-TARS-72B          & 18.8 & 12.5 & 62.9 & 17.2 & 57.1 & 15.4 & 64.6 & 20.9 & 63.3 & 26.4 & 42.1 & 15.7 & 50.9 & 17.6 & 38.1\\
        JEDI-3B              & 27.4 & 9.4 & 61.0 & 13.8 & 53.5 & 8.4 & 54.2 & 18.2 & 64.4 & 32.1 & 38.3 & 9.0 & 49.8 & 13.7 & 36.1\\
        JEDI-7B              & 38.0 & 14.1 & 42.9 & 11.0 & 50.0 & 11.9 & 72.9 & 25.5 & 75.1 & 47.2 & 33.6 & 16.9 & 52.6 & 18.2 & 39.5\\
        GUI-Actor-7B         & – & – & – & – & – & – & – & – & – & – & – & – & – & – & 44.6\\
        \midrule
        \multicolumn{16}{c}{\textbf{\textit{GUI-specific Models (RL)}}}\\
        \addlinespace
        UI-R1-3B             & 11.2 & 6.3 & 22.7 & 4.1 & 27.3 & 3.5 & 42.4 & 11.8 & 32.2 & 11.3 & 13.1 & 4.5 & 24.9 & 6.4 & 17.8\\
        UI-R1-E-3B           & 37.1 & 12.5 & 46.1 & 6.9 & 41.9 & 4.2 & 56.9 & 21.8 & 65.0 & 26.4 & 32.7 & 10.1 & – & – & 33.5\\
        GUI-R1-3B            & 26.4 & 7.8 & 33.8 & 4.8 & 40.9 & 5.6 & 61.8 & 17.3 & 53.6 & 17.0 & 28.1 & 5.6 & – & – & –\\
        GUI-R1-7B            & 23.9 & 6.3 & 49.4 & 4.8 & 38.9 & 8.4 & 55.6 & 11.8 & 58.7 & 26.4 & 42.1 & 16.9 & – & – & –\\
        InfiGUI-R1-3B        & 33.0 & 14.1 & 51.3 & 12.4 & 44.9 & 7.0 & 58.3 & 20.0 & 65.5 & 28.3 & 43.9 & 12.4 & 49.1 & 14.1 & 35.7\\
        GUI-G1-3B            & 39.6 & 9.4 & 50.7 & 10.3 & 36.6 & 11.9 & 61.8 & 30.0 & 67.2 & 32.1 & 23.5 & 10.6 & 49.5 & 16.8 & 37.1\\
        SE-GUI-3B            & 38.1 & 12.5 & 55.8 & 7.6 & 47.0 & 4.9 & 61.8 & 16.4 & 59.9 & 24.5 & 40.2 & 12.4 & 50.4 & 11.8 & 35.9\\
        SE-GUI-7B            & 51.3 & 42.2 & 68.2 & 19.3 & 57.6 & 9.1 & 75.0 & 28.2 & 78.5 & 43.4 & 49.5 & 25.8 & 63.5 & 21.0 & 47.3\\
        GUI-$\text{G}^2$-7B         & 55.8 & 12.5 & 68.8 & 17.2 & 57.1 & 15.4 & 77.1 & 24.5 & 74.0 & 32.7 & 57.9 & 21.3 & 64.7 & 19.6 & 47.5\\
        \midrule
        \multicolumn{16}{c}{\textbf{\textit{Ours}}}\\
        \addlinespace
        \textbf{V2P-7B}     & \textbf{58.38}	& 12.50 & 67.53	& \textbf{24.83} & \textbf{62.63}	& \textbf{16.08} & 73.61	& \textbf{33.64} & 75.71	& 43.40 & 56.07	& \textbf{32.58} & \textbf{65.81}	& \textbf{25.83}	& \textbf{52.50}\\
        \bottomrule
        \end{tabular}
    }
    \caption{Comparison of Model Performance Across Task Categories in ScreenSpot-Pro. Bold text highlights the best results, while ``–'' represents missing values not reported in the original papers. The baseline models utilize various backbones and parameter sizes, as indicated by their names (e.g., -7B, -18B).}
    \label{tab:screenspot_pro}
\end{table*}
\vspace{-0.2cm}

\section{Qualitative Analysis and Case Studies}
\label{app:qualitative_analysis}

\subsection{Success Cases}

\begin{figure*}[htbp]
    \centering

    \begin{subfigure}[b]{0.7\textwidth}
        \centering
        \includegraphics[width=\textwidth]{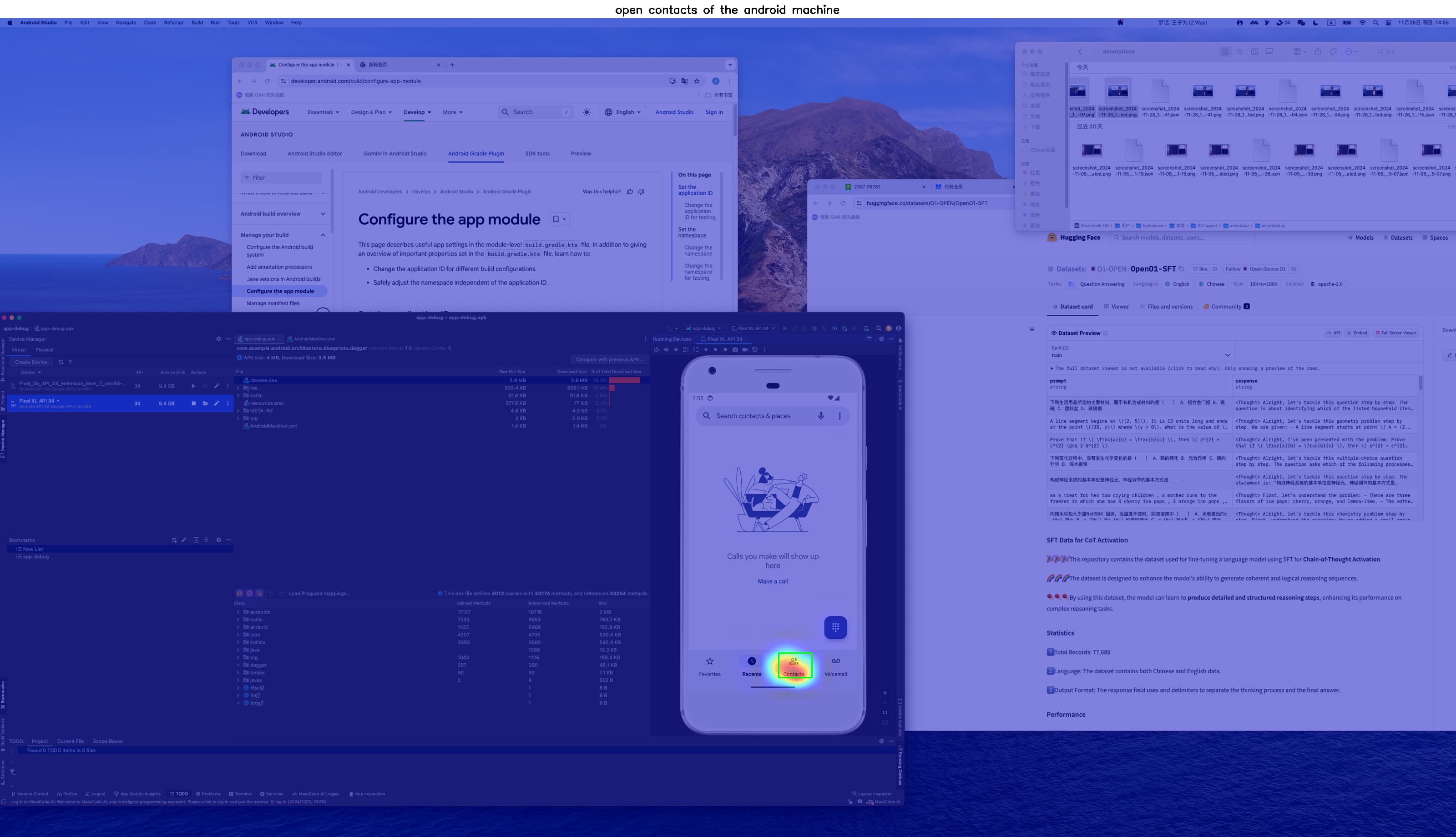}
        \caption{Success Case 1}
        \label{fig:success_case1}
    \end{subfigure}

    \vspace{0.5em} 

    \begin{subfigure}[b]{0.7\textwidth}
        \centering
        \includegraphics[width=\textwidth]{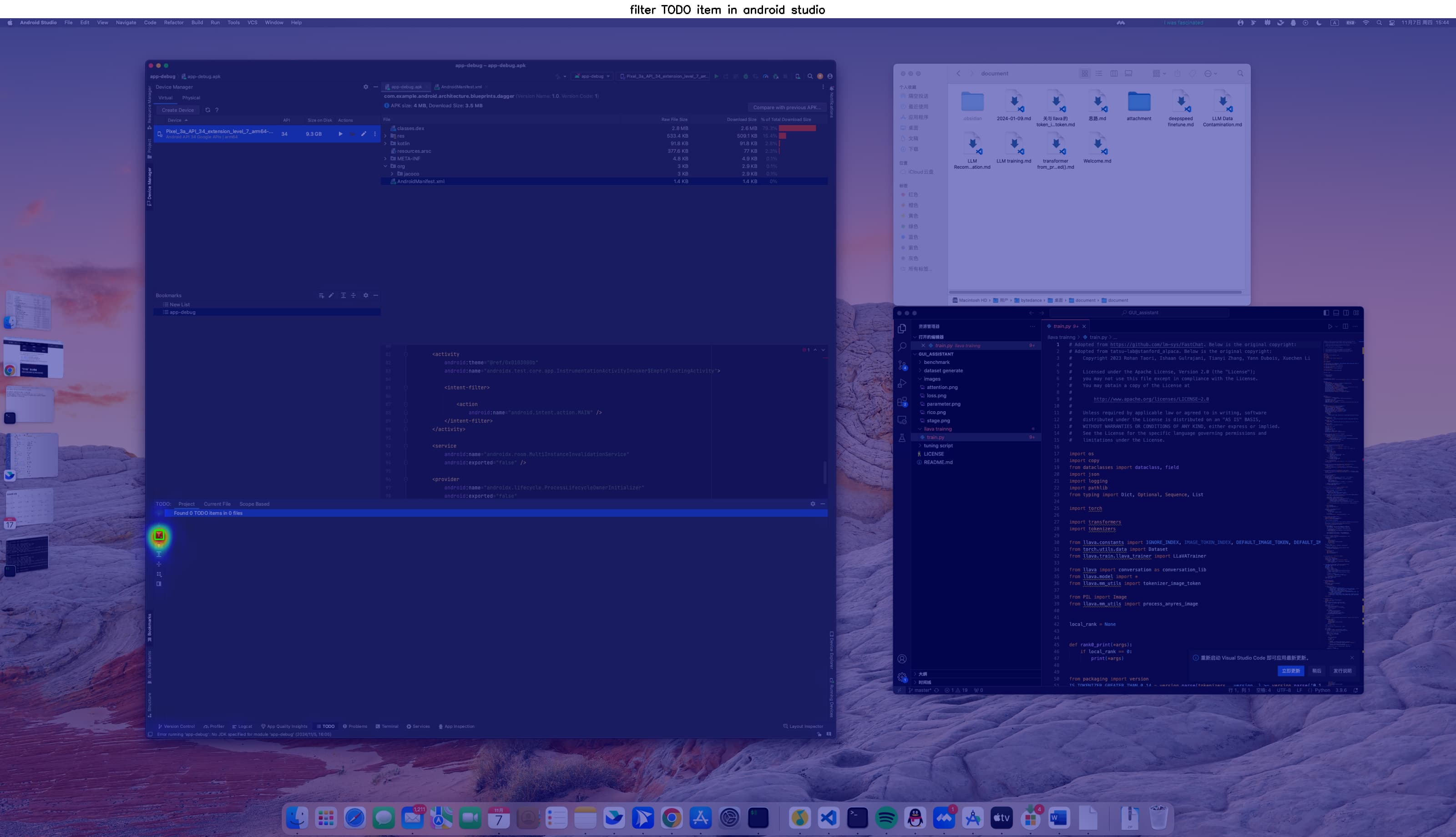}
        \caption{Success Case 2}
        \label{fig:success_case2}
    \end{subfigure}

    \vspace{0.5em} 

    \begin{subfigure}[b]{0.7\textwidth}
        \centering
        \includegraphics[width=\textwidth]{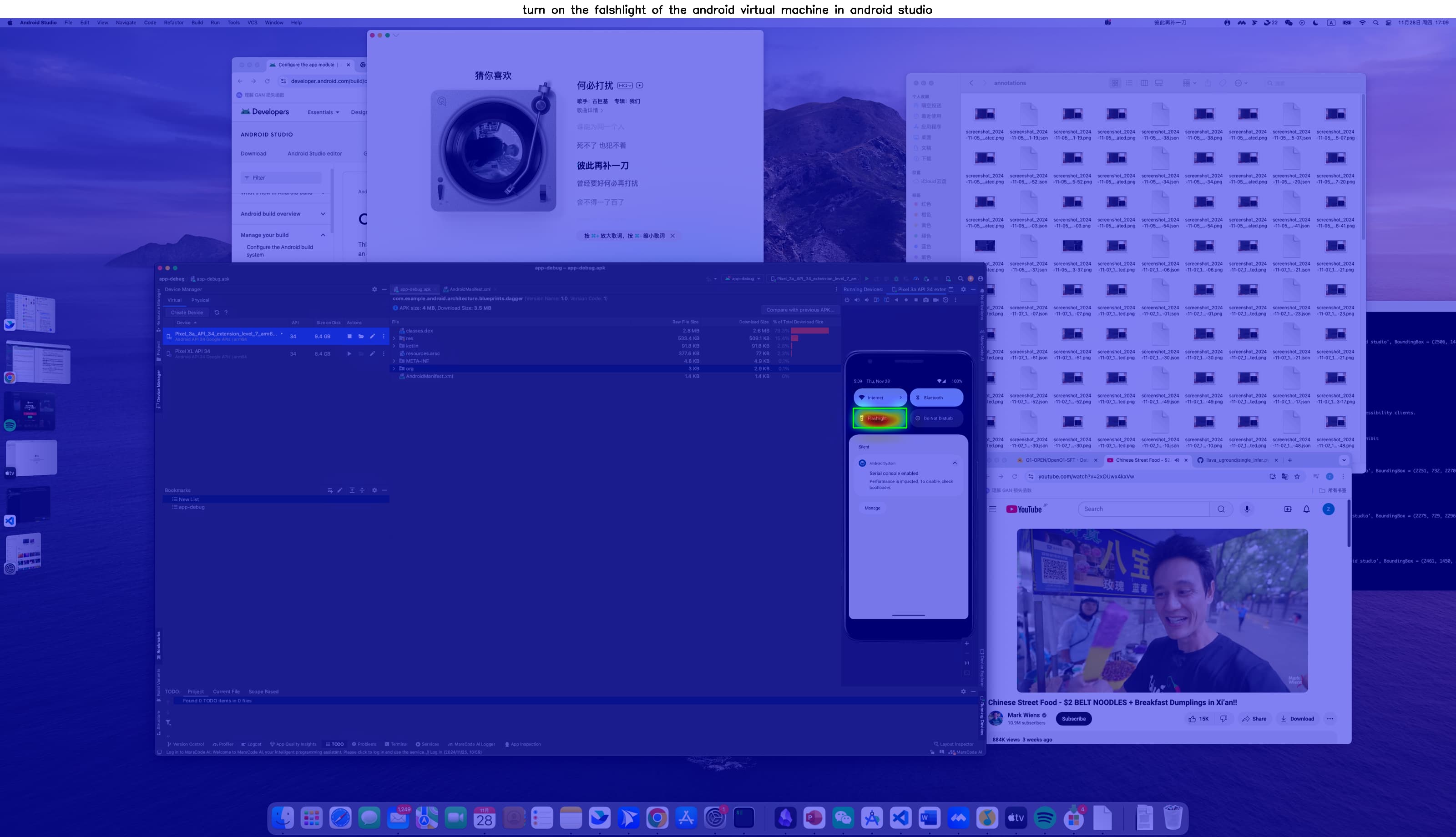}
        \caption{Success Case 3}
        \label{fig:success_case3}
    \end{subfigure}

    \caption{Representative success cases of GUI element localization.}
    \label{fig:success_cases}
\end{figure*}

Fig.~\ref{fig:success_cases} demonstrate several representative success cases where our V2P-7B model achieves accurate GUI element localization. Through these successful examples, we observe that the model exhibits high confidence in precisely highlighting target regions, with attention distributions that closely align with the actual shapes of UI elements. The attention maps show sharp, well-defined boundaries that accurately correspond to button edges, text field borders, and icon contours. This demonstrates the model's robust understanding of visual-semantic correspondence between natural language instructions and GUI components, effectively bridging the gap between textual descriptions and visual interface elements.

\subsection{Failure Cases and Error Analysis}

\begin{figure*}[htbp]
    \centering

    \begin{subfigure}[b]{0.7\textwidth}
        \centering
        \includegraphics[width=\textwidth]{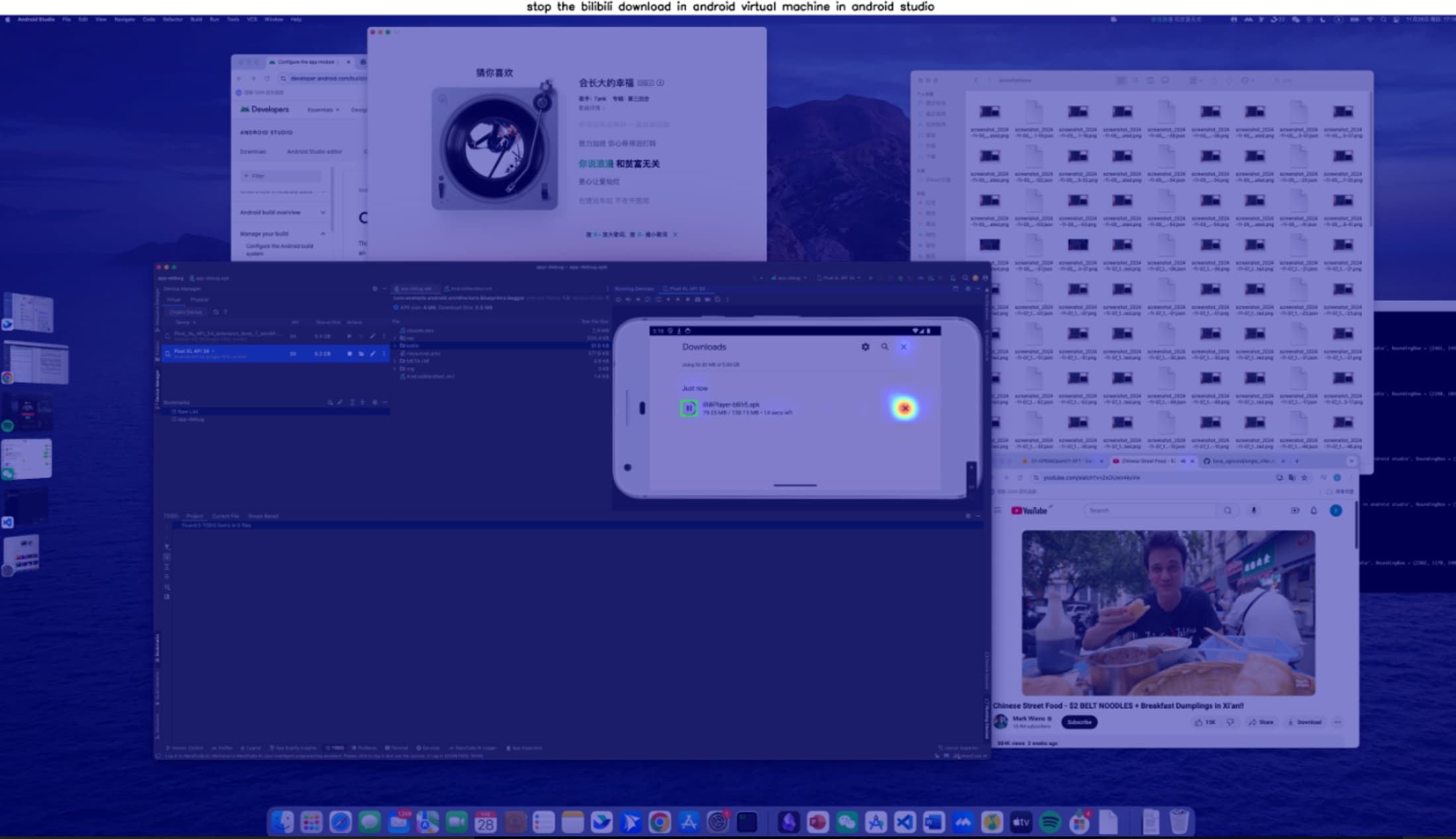}
        \caption{Failure Case 1}
        \label{fig:failure_case1}
    \end{subfigure}

    \vspace{0.5em} 

    \begin{subfigure}[b]{0.7\textwidth}
        \centering
        \includegraphics[width=\textwidth]{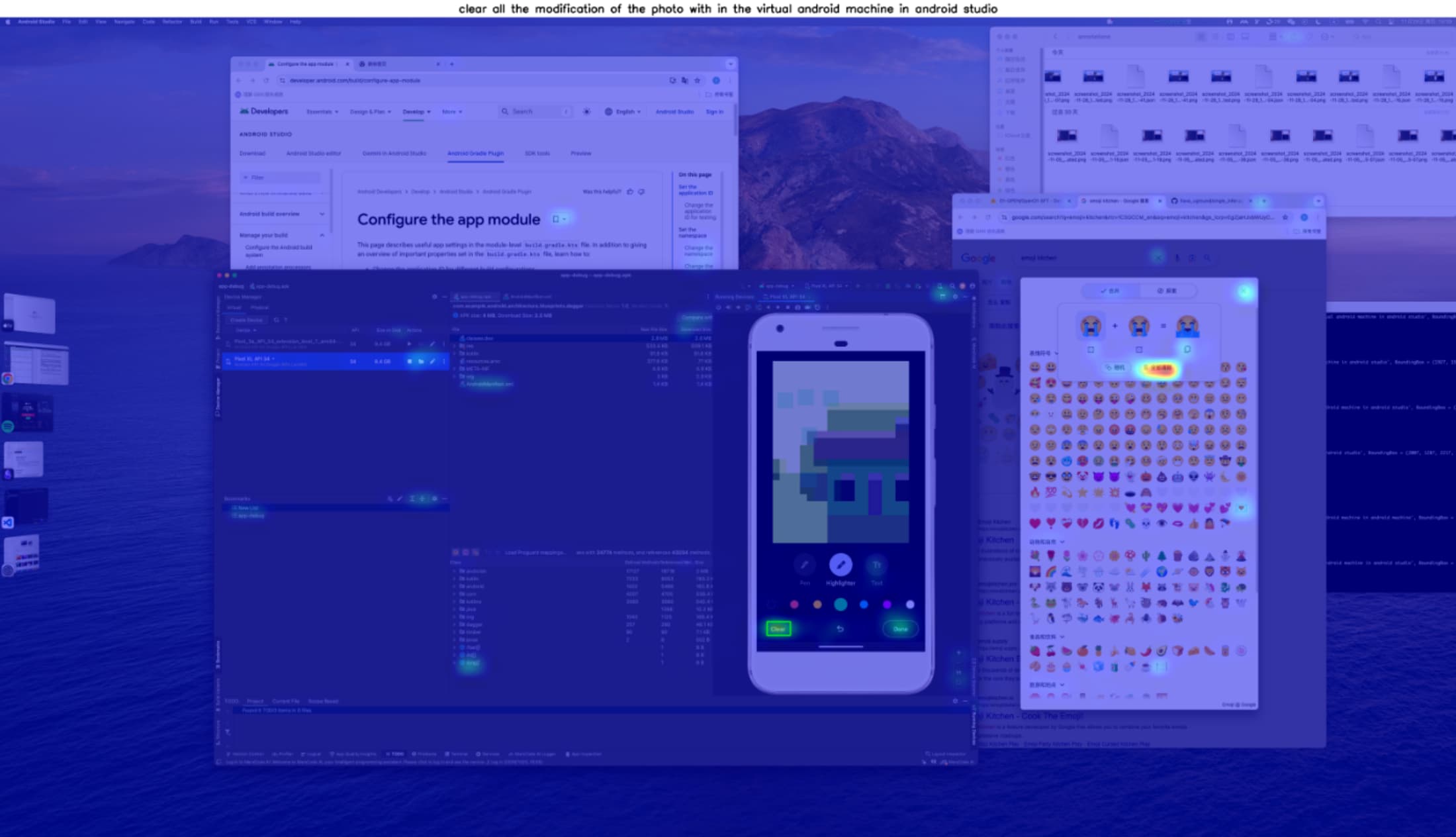}
        \caption{Failure Case 2}
        \label{fig:failure_case2}
    \end{subfigure}

    \vspace{0.5em} 

    \begin{subfigure}[b]{0.7\textwidth}
        \centering
        \includegraphics[width=\textwidth]{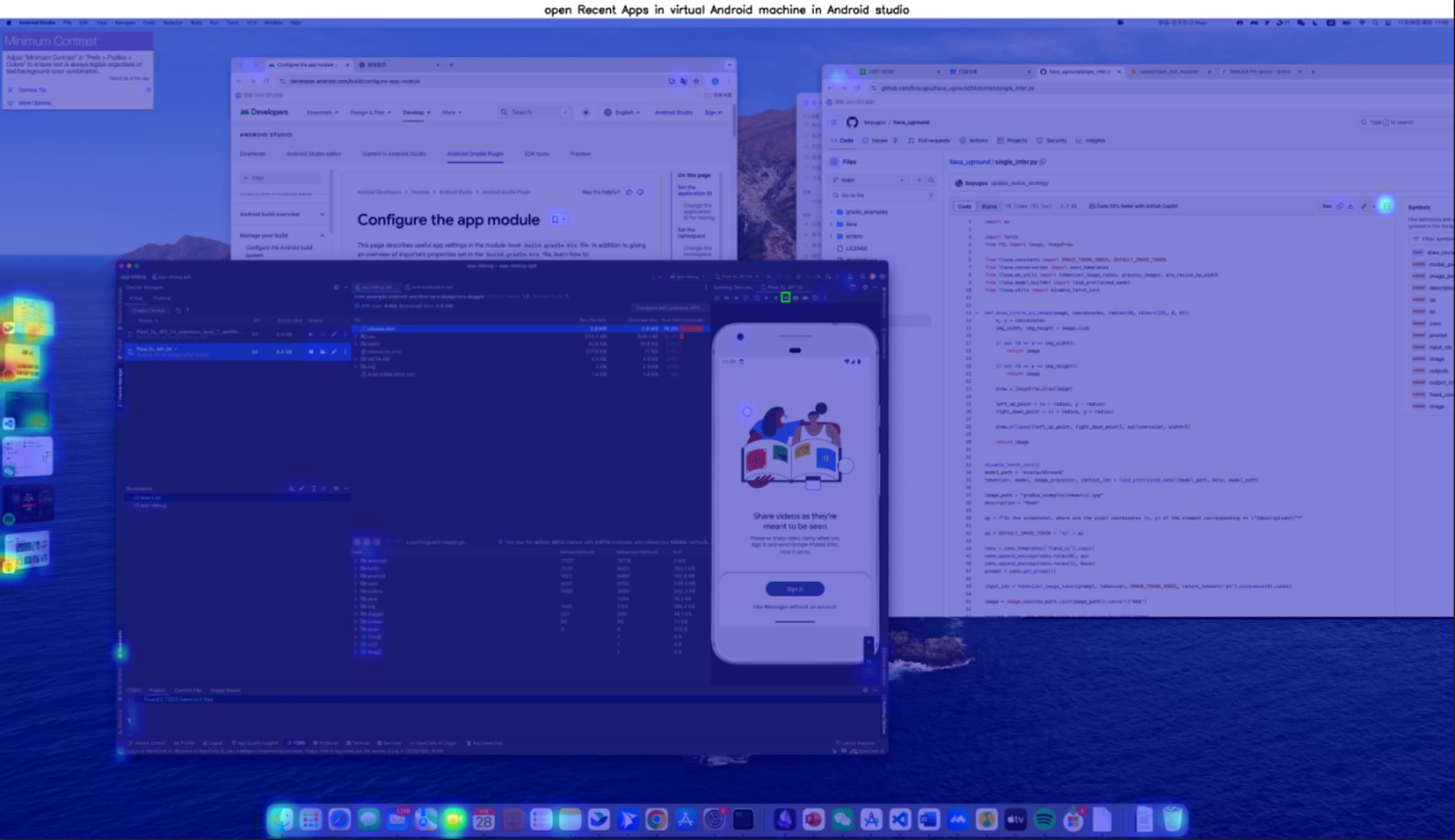}
        \caption{Failure Case 3}
        \label{fig:failure_case3}
    \end{subfigure}

    \caption{Representative failure cases of GUI element localization.}
    \label{fig:failure_cases}
\end{figure*}

Our analysis of failure cases reveals several interesting patterns and limitations, as illustrated in Fig.~\ref{fig:failure_cases}. In some instances, we observe that the model encounters difficulties when multiple UI elements share semantic similarities. The model often exhibits high confidence while incorrectly selecting semantically related but functionally different elements or misidentifying similar icons with different purposes (Fig.~\ref{fig:failure_case1}).

Additionally, we identify cases where the model's attention distribution becomes highly dispersed across the interface, which we interpret as an indicator of \textit{low confidence} (Fig.~\ref{fig:failure_case2}). This scattered attention pattern typically occurs in scenarios with numerous distracting elements or cluttered interfaces, suggesting that the model's decision-making process becomes uncertain when faced with complex visual layouts.

Furthermore, we observe failure modes where the model's attention concentrates entirely on regions completely unrelated to the target element (Fig.~\ref{fig:failure_case3}). These cases often involve ambiguous natural language descriptions or interfaces with unconventional design patterns that deviate from the model's training distribution. Such failures highlight the need for enhanced user intent understanding and more comprehensive UI context comprehension capabilities.

\subsection{Multi-step Interaction Scenarios}

\begin{figure*}[htbp]
    \centering
    \begin{subfigure}[b]{0.3\textwidth}
        \centering
        \includegraphics[width=\textwidth]{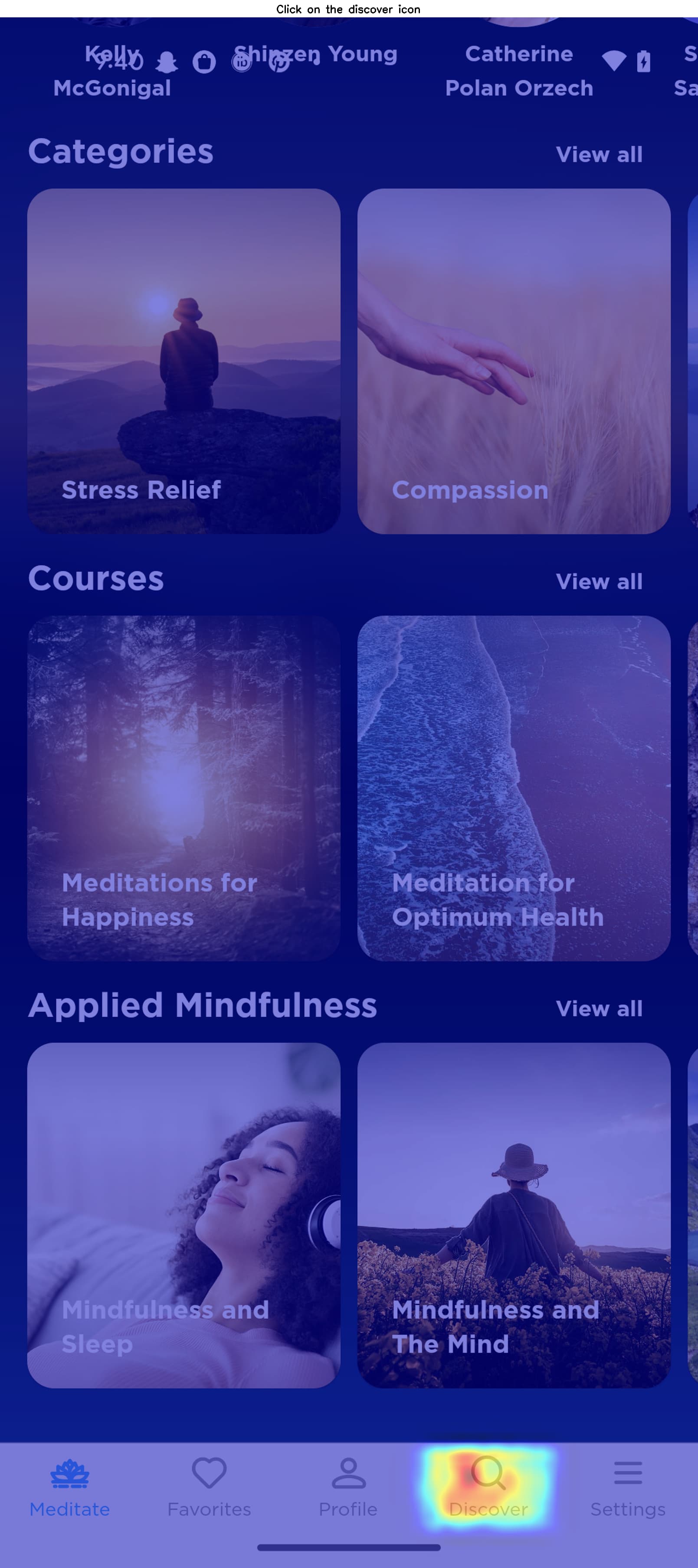}
        \caption{Step 1: Click on the discover icon.}
        \label{fig:multi_step_case2_step1}
    \end{subfigure}
    \hfill
    \begin{subfigure}[b]{0.3\textwidth}
        \centering
        \includegraphics[width=\textwidth]{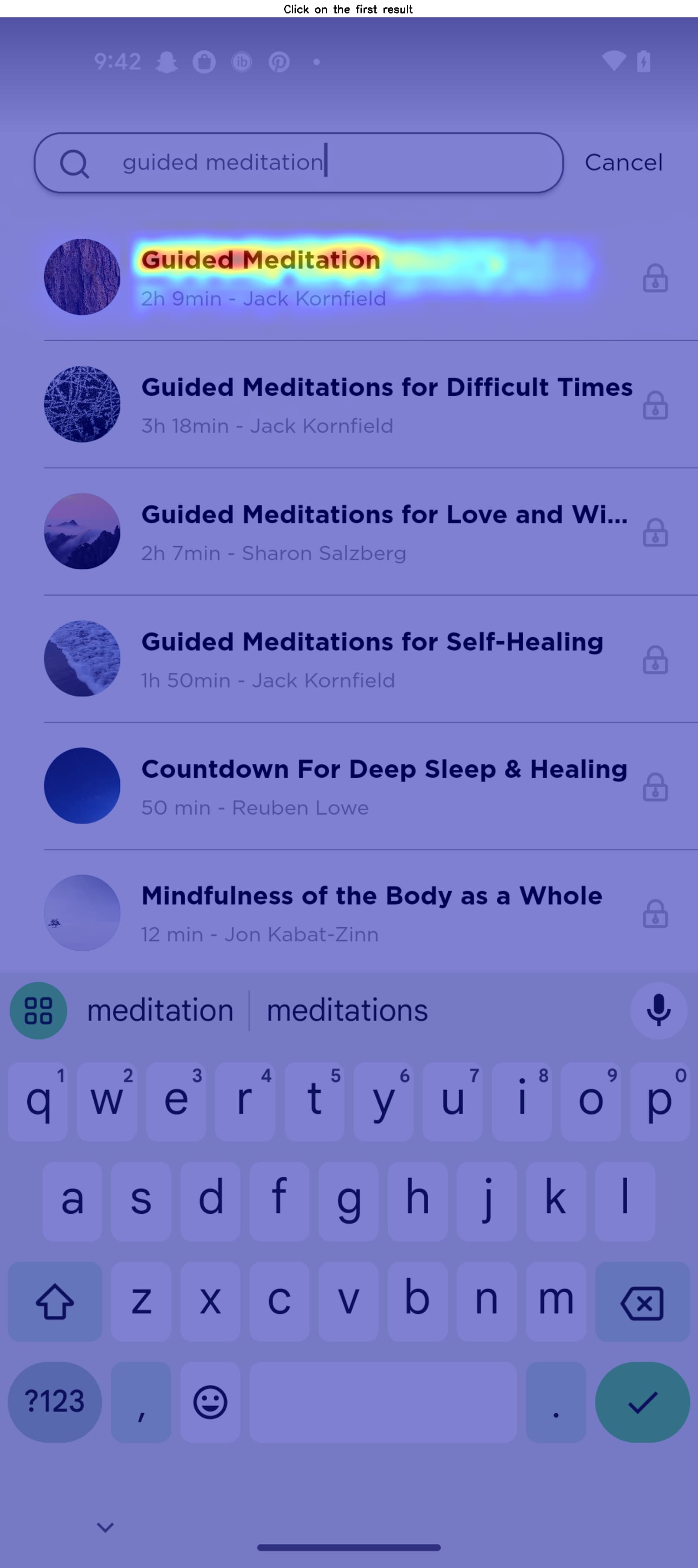}
        \caption{Step 2: Click on the first result.}
        \label{fig:multi_step_case2_step2}
    \end{subfigure}
    \hfill
    \begin{subfigure}[b]{0.3\textwidth}
        \centering
        \includegraphics[width=\textwidth]{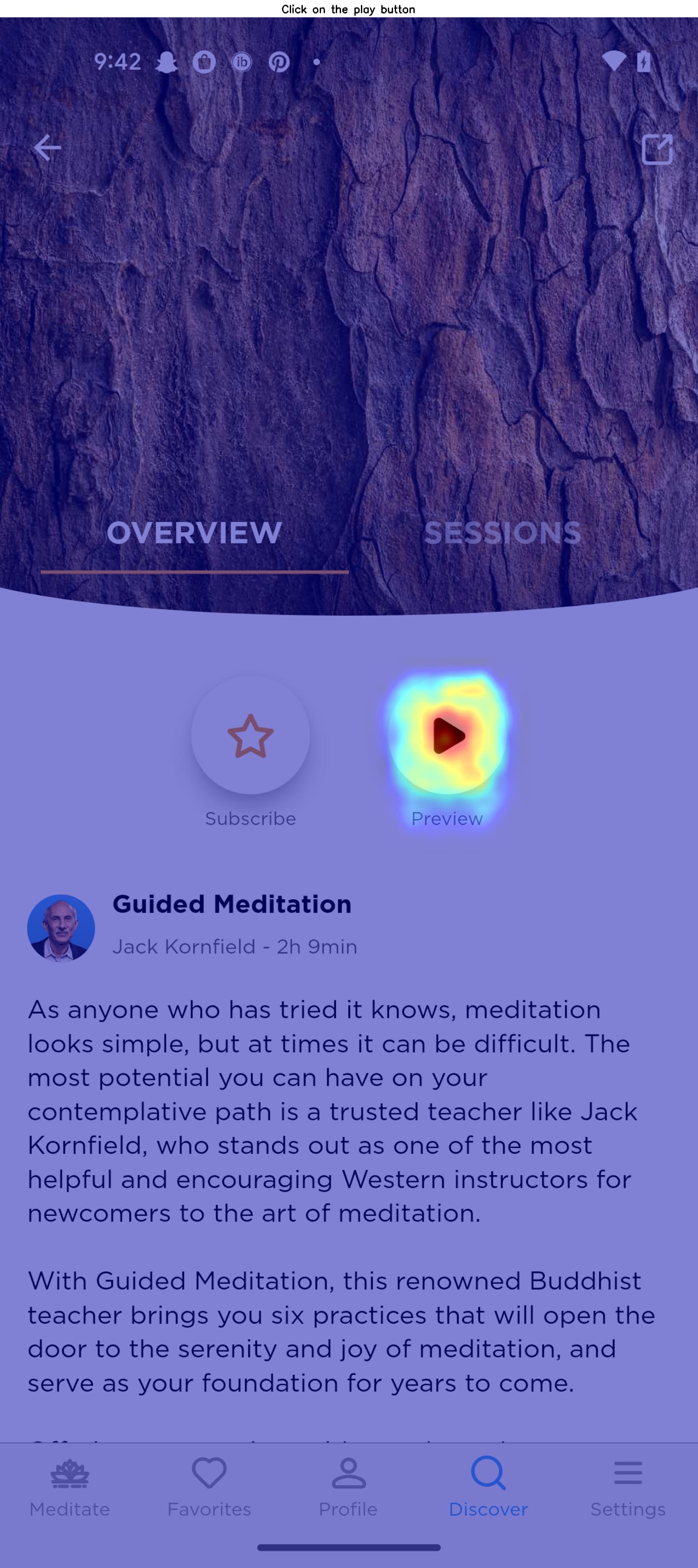}
        \caption{Step 3: Click on the play button.}
        \label{fig:multi_step_case2_step3}
    \end{subfigure}
    \caption{Multi step grounding case 2: "Open the Mindfulness app, I would like to have a personalized guided meditation to help me be productive throughout the day."}
    \label{fig:multi_step_case2}
\end{figure*}

To visualize the model's capability in maintaining context across sequential operations, we present case studies of multi-step workflows from the AndroidControl~\citep{55} dataset. Fig.~\ref{fig:multi_step_case1} and Fig.~\ref{fig:multi_step_case2} showcases the model's performance across sequential GUI operations.

The results demonstrate that our model maintains consistent accuracy throughout extended interaction sequences, successfully completing multi-step tasks that require contextual understanding and state awareness.

\subsection{Multi-target Localization Capabilities}

\begin{figure*}[t]
    \centering

    \begin{subfigure}[t]{0.61\textwidth}
        \centering
        \begin{minipage}[t]{0.49\linewidth}
            \centering
            \includegraphics[width=\textwidth]{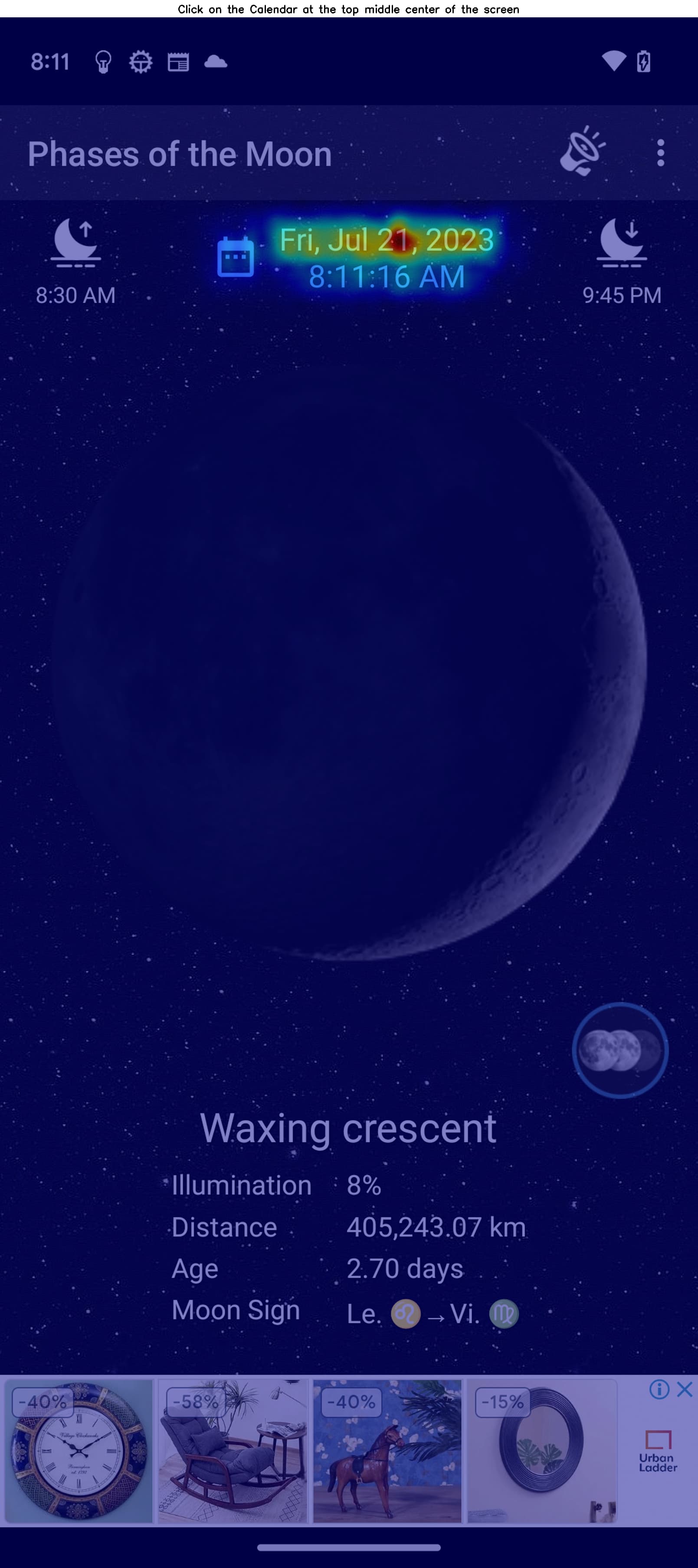}
        \end{minipage}
        \hfill
        \begin{minipage}[t]{0.49\linewidth}
            \centering
            \includegraphics[width=\textwidth]{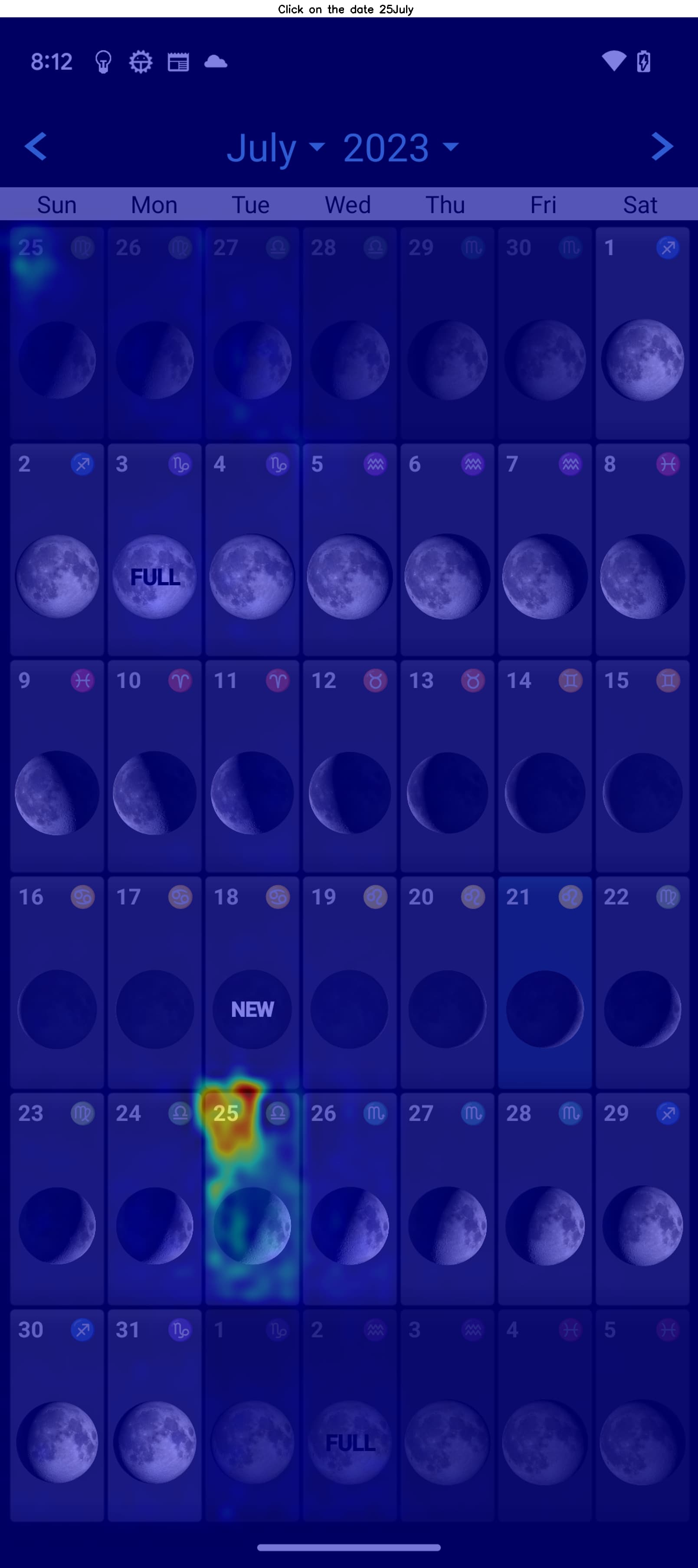}
        \end{minipage}
        
        \caption{Multi step grounding case 1: "Open Phase of the moon App, select the date 25 July on the calendar and view the moon phase for that date." Step 1 (left) and Step 2 (right).}
        \label{fig:multi_step_case1}
    \end{subfigure}
    \hfill 
    \begin{subfigure}[t]{0.3\textwidth}
        \centering
        \includegraphics[width=\linewidth]{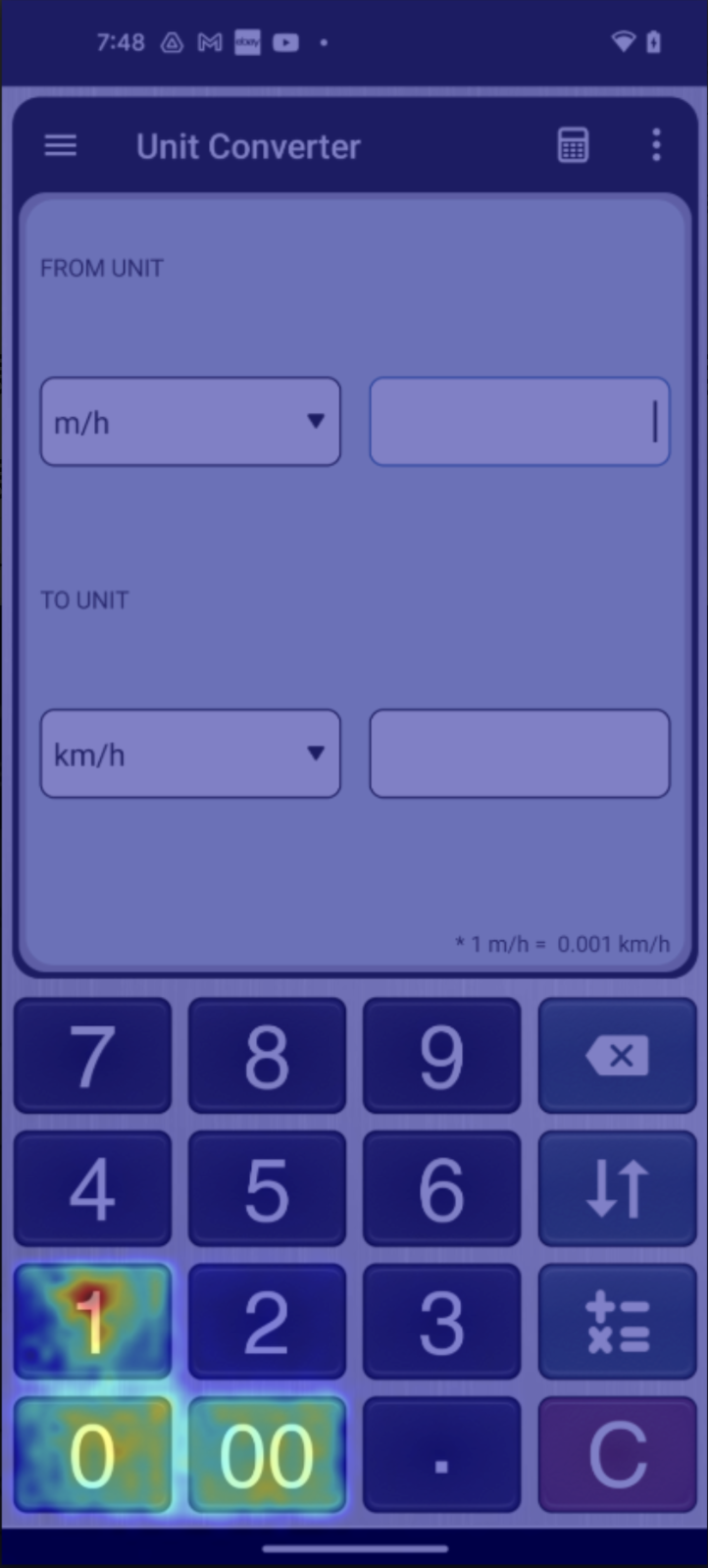}
        \caption{Multi-target grounding case.}
        \label{fig:multitarget_localization}
    \end{subfigure}
    
    \caption{Multi-step grounding case and multi-target grounding case.}
    \label{fig:combined_cases}
\end{figure*}

We investigated the model's ability to simultaneously localize multiple targets within a single interface, which holds significant value for batch operations and improving inference efficiency. Fig.~\ref{fig:multitarget_localization} presents our experimental setup using a calculator interface, where we tasked the model with simultaneously localizing the elements "1", "0", and "00".

The results reveal that the model successfully generates attention distributions for all three target elements simultaneously, with appropriately differentiated confidence levels. Notably, the element "1" receives the highest attention intensity, followed by "0" and "00" respectively, which aligns with the natural priority of these elements. This multi-target capability demonstrates the model's sophisticated attention mechanism and its potential for complex GUI analysis tasks requiring simultaneous element identification, as well as its genuine understanding capability of user queries.

\label{sec:appendix}
\end{document}